%% file: acl2024.tex
\pdfoutput=1

\documentclass[11pt]{article}

\usepackage[]{ACL2023}

\usepackage{times}
\usepackage{latexsym}

\usepackage[T1]{fontenc}

\usepackage[utf8]{inputenc}

\usepackage{microtype}
\input{config}

\usepackage{inconsolata}

\usepackage{graphicx}
\usepackage{booktabs}

\newcommand{\task}{TL~}
\title{A Law Reasoning Benchmark for LLM with Tree-Organized Structures including Factum Probandum, Evidence and Experiences}

\author{
Jiaxin Shen \textsuperscript{1}, \quad  Jinan Xu \textsuperscript{1}, \quad Huiqi Hu \textsuperscript{1}, \quad Luyi Lin \textsuperscript{1}, \quad Fei Zheng \textsuperscript{2} \\
Guoyang Ma \textsuperscript{2}, \quad Fandong Meng \textsuperscript{3}, \quad Jie Zhou \textsuperscript{3}, \quad Wenjuan Han\thanks{\quad Wenjuan Han is the corresponding author} \ \textsuperscript{1} \\
\small1. Beijing Key Laboratory of Traffic Data Mining and Embodied Intelligence, Beijing, China; \\ 
\small2. Law School, Beijing Jiaotong  University, Beijing, China; \\
\small3. Pattern Recognition Center, WeChat AI, Tencent Inc, China\\
\small\texttt{\{jxshen, jaxu, huiqi\_hu, luyilin, gyma1, wjhan\}@bjtu.edu.cn} \\
\small\texttt{\{fandongmeng, withtomzhou\}@tencent.com} \\
}

\begin{document}
\maketitle
\begin{abstract}
While progress has been made in legal applications, law reasoning, crucial for fair adjudication, remains unexplored. We propose a transparent law reasoning schema enriched with hierarchical factum probandum, evidence, and implicit experience, enabling public scrutiny and preventing bias. Inspired by this schema, we introduce the challenging task, which takes a textual case description and outputs a hierarchical structure justifying the final decision. We also create the first crowd-sourced dataset for this task, enabling comprehensive evaluation. Simultaneously, we propose an agent framework that employs a comprehensive suite of legal analysis tools to address the challenge task. This benchmark paves the way for transparent and accountable AI-assisted law reasoning in the ``Intelligent Court''\footnote{The code and data are available at \url{https://github.com/cocacola-lab/LawReasoningBenchmark}
}.
\end{abstract}

\section{Introduction}\label{sec:intro}

In recent times, Artificial Intelligence (AI) has demonstrated a profound impact on legal applications, including the generation of legal document summarization~\citep{jain2023sentence}, argument mining,~\citep{Xu2021summarizing} and legal case retrieval~\citep{Ma2023Structural, liu2023leveraging}. 
While recent advances focus on generating impartial and interpretable judicial judgments based on established criminal fact~\citep{t-y-s-s-etal-2024-towards, he-etal-2024-agentscourt, HAN2024121052}. However, the premise for ensuring this process is the accurate determination of the ultimate criminal facts. The fundamental challenge remains: how to construct logically rigorous, evidence-backed ultimate criminal facts from evidentiary materials and inferred interim facts.

Accurate criminal fact determination forms the cornerstone of judicial fairness~\citet{ allen2010evidence, anderson2005analysis,chafee1931principles}. However, existing AI judges primarily address post-fact legal procedures rather than simulating comprehensive court processes.  the fairness of adjudication fundamentally depends on systematic evidence analysis and fact reasoning during fact-finding phases. Therefore, we shift focus to an underexplored frontier: \textbf{Law Reasoning}\footnote{Law Reasoning is also known as evidence reasoning and evidence analysis.}, aiming to bridge the gap between evidence interpretation and judicial decision-making.

\begin{figure}[t!]
    \centering
    \includegraphics[width=\linewidth]{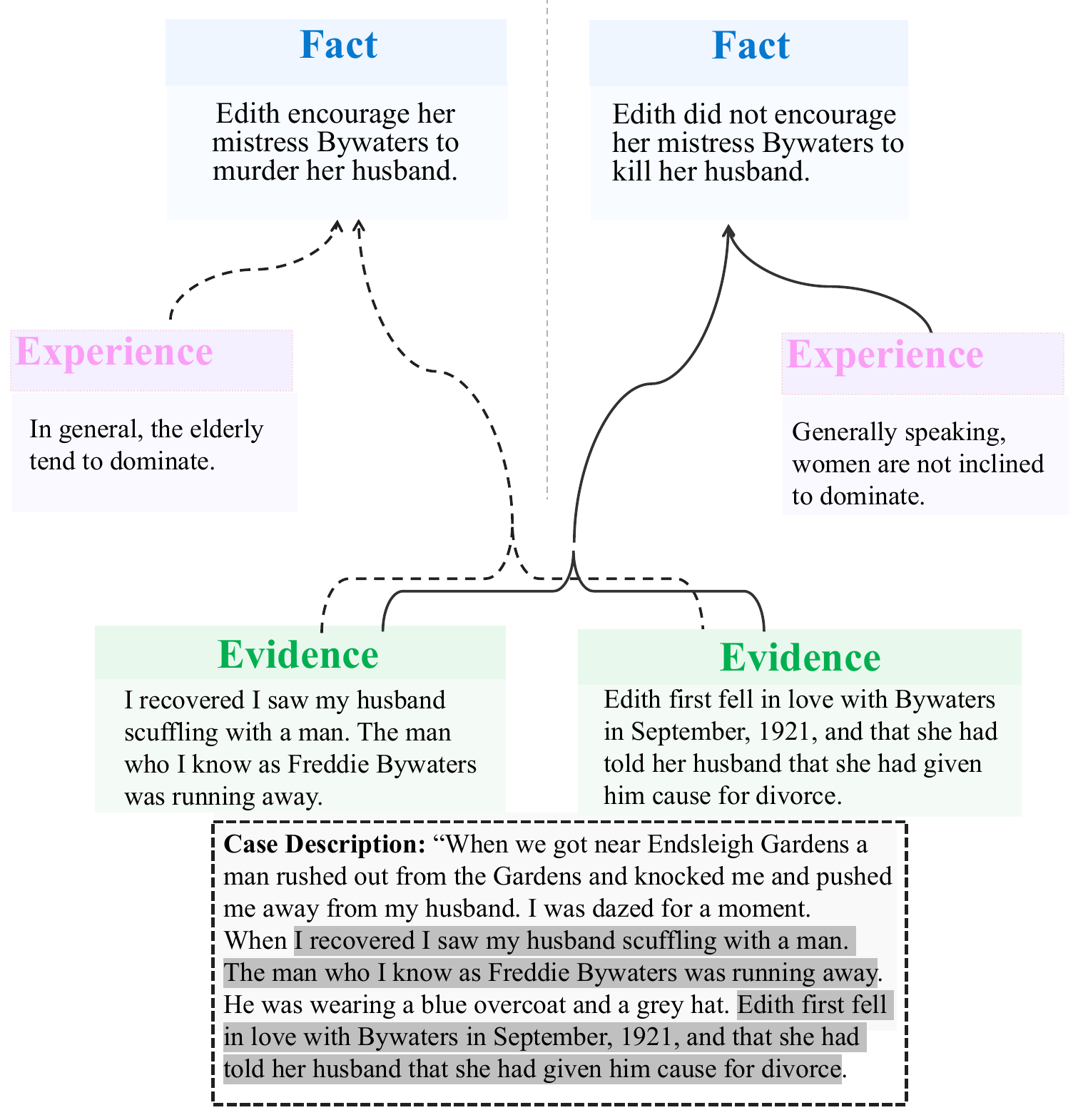}
    \caption{Case ``Rex v. Bywaters and Thompson'' that demonstrates different experiences have impacted different results (LEFT vs. RIGHT). The case description and evidence are shared, but the experiences of both sides are different, which leads to different ultimate probandum.}
    \label{fig:motivation}
    \vspace{-5mm}
\end{figure}



To highlight the significance of Law Reasoning, we provide examples that are widely recognized where different evidence and human experience lead to different criminal facts. Recognizing these instances is crucial for maintaining judicial justice and public trust. Here is a notable example in Figure \ref{fig:motivation}.

In these cases and many others before them, it is evident that wrongful judgments often arise due to the misuse of experience.
To mitigate this risk, we aim to make the law reasoning procedure \textbf{transparent} and make the details visible through the law reasoning process employed by judges to subject judicial activities to social supervision, prevent the influence of prejudice, promote social fairness and justice, and enhance public trust in the judiciary.

In light of the essential of transparent law reasoning, a schema that accurately simulates law reasoning process is desired. \citet{wigmore1937science} has long proposed a diagram method for law reasoning. However, these diagram methods remain at a theoretical level due to their complex structure and numerous elements. To address this, \citet{anderson2005analysis} have enhanced Wigmore's diagrammatic method to make it more user-friendly.
Taking inspiration from these iconographic methods, we have adopted a modified version that enriches the schema by incorporating implicit experience. The modified schema shows a justification procedure of facts made by the fact finder (Jury or judge) at a trial.
Section \ref{sec:schema_formulation} provides a visual representation and detailed explanation of the schema. 

Then with the designed schema as a foundation, we introduce a new challenging task --- \textbf{T}ransparent \textbf{L}aw Reasoning with Tree-Organized Structures  (\task for short), which aims to generate the hierarchical and layered law reasoning structure following the schema (for ease of understanding, we explain the legal terms involved in \textbf{Table \ref{tab:glossary}} and use them later to describe our work). In this challenge, the textual case description is input, and the \task task is to output the law reasoning structure where the top node represents the terminal fact. Specifically, we formalize the \task procedure as a tree-like structure. Each node involves a step of reasoning from evidence to interim probandum that need to be proven, and then from interim probandum to ultimate probandum. 
Additionally, we conduct the first comprehensive and qualitative study on law reasoning simulation at a trial by introducing a crowd-sourcing development/test dataset for evaluation (Section \ref{sec:data_collection}). 


In summary, our contributions are three-fold: (i) A schema enhanced with hierarchical factum probandum\footnote{The factum probandum (pl. facta probanda) refers to the fact to be proved.}, evidence, and experiences fusion; (ii) A new challenging task – \task with crowdsourcing data and corresponding metrics; (iii) The \task agent utilizes a comprehensive suite of legal analysis tools to construct tree-organized structures.

\section{Task Definition}\label{sec:task}
\begin{figure}[t!]
    \centering
    \includegraphics[width=0.7\linewidth]{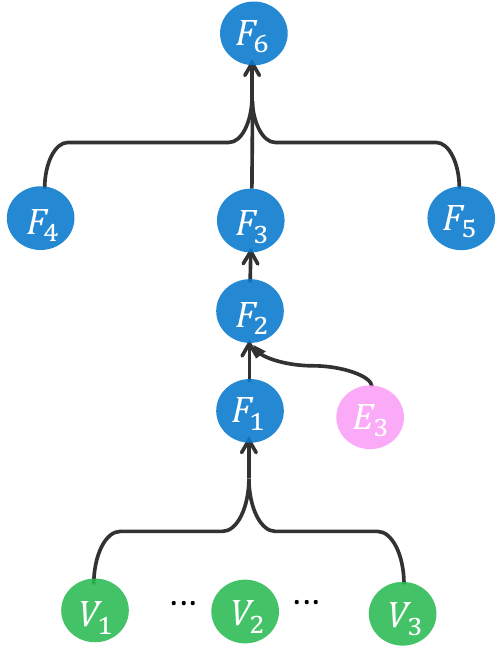}
    \caption{Illustration of the schema.}
    \label{fig:schema}
    \vspace{-5mm}
\end{figure}

\begin{figure*}[htp]
\centering
\includegraphics[width=1.0\linewidth]
{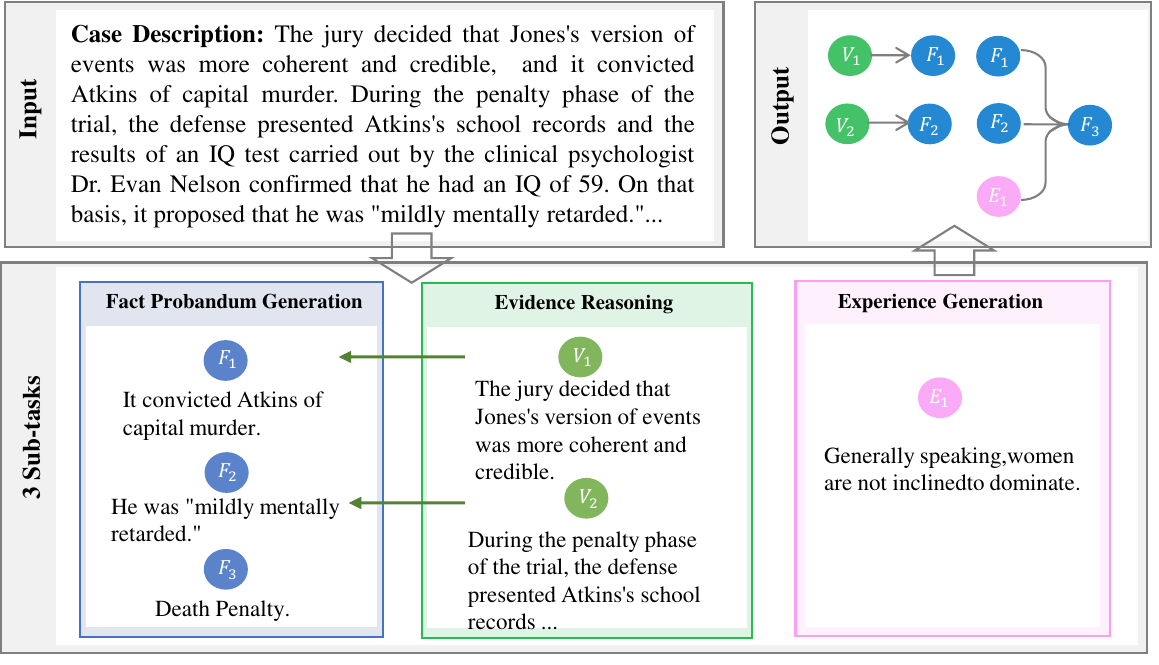}
\caption{Illustration of the task. For convenience, we showcase examples for each sub-task. The output of the 3 sub-tasks is collected to form the complete law reasoning structure.}
\label{fig:task}
\end{figure*}

We start by briefly introducing our schema formulation in Section \ref{sec:schema_formulation}. Then we present our task \task from task formulation (Section \ref{sec:task_formulation}) to metrics (Section \ref{sec:metrics} in appendix) in detail.

\subsection{Schema Formulation}\label{sec:schema_formulation}
The term ``law reasoning'' is used to describe the process of reasoning from the case description $\mathbf{x}$ to the ultimate probandum to be proved, which determines the inductive logical nature of judicial proof. The designed schema rigorously represent this process, showing how ultimate probandum are finally obtained from original evidences. The process starts from evidences, goes through different granularity of factum probandum, and gets the ultimate probandum.
We introduce the schema formulation in Figure \ref{fig:schema}. To make the implicit law reasoning process transparent, all factum probandum, evidence, experiences and the supportive relationships between them need to be revealed.
The schema is formed in a nested tree similar to \citet{anderson2005analysis}, including the following four elements:

\noindent\textbullet~\textbf{Evidence.} Following the real judicial process, the basic e\textbf{V}idence block $\mathbf{V}$ of the schema includes documentary evidences, testimonial evidences, among others. The evidence node is the leaf node from which legal practitioners or intelligent models need to infer that certain factum probandum did occur. $\mathrm{v}_1, \mathrm{v}_2, \mathrm{v}_3 \in \mathbf{V}$.
\\
\noindent\textbullet~\textbf{Factum Probandum.} Factum probandum have multiple levels of granularity, including interim probandum ($\mathrm{f}_1, \mathrm{f}_2$), penultimate probandas ($\mathrm{f}_3, \mathrm{f}_4, \mathrm{f}_5$), and ultimate probandum ($\mathrm{f}_6$), from fine to coarse. More coarse ones are made up of fine ones.  Fine-to-coarse factum probandum $[\mathrm{f}_1, \mathrm{f}_2, \mathrm{f}_3, \mathrm{f}_4, \mathrm{f}_5]$  guide a series of inference connecting the evidences $[\mathrm{v}_1, \mathrm{v}_2, \mathrm{v}_3..., \mathrm{v}_n]$ with the ultimate probandum $\mathrm{f}_6$.$\mathrm{f_i} \in \mathrm{F}$.
\footnote{Facts are renamed as propositions in some legal articles.}

\noindent\textbullet~\textbf{Experiences.} Human Experience $\mathrm{e}$ used during connecting evidence $\mathrm{v}$ and fact $\mathrm{f}$, and forming coarse factum probandum. Practitioners or intelligent models may need personal experiences for reasoning. The experiences help to explain why the decision maker inference like this, making the process more explicit to understand. 

\noindent\textbullet~\textbf{Inferences.} The edges $\mathrm{r}$ in the reasoning process, support each reasoning step and authorize the inference from the bottom up.  Inferences exist between evidences $\mathrm{v}$ and factum probandum $\mathrm{f}$, as well as between different granularity of factum probandum. Formally, $\mathrm{r:v\longrightarrow f}$ under $\mathrm{e}$.



\subsection{Task Formulation}\label{sec:task_formulation}
\begin{figure}[h]
\centering
\includegraphics[width=1\linewidth]{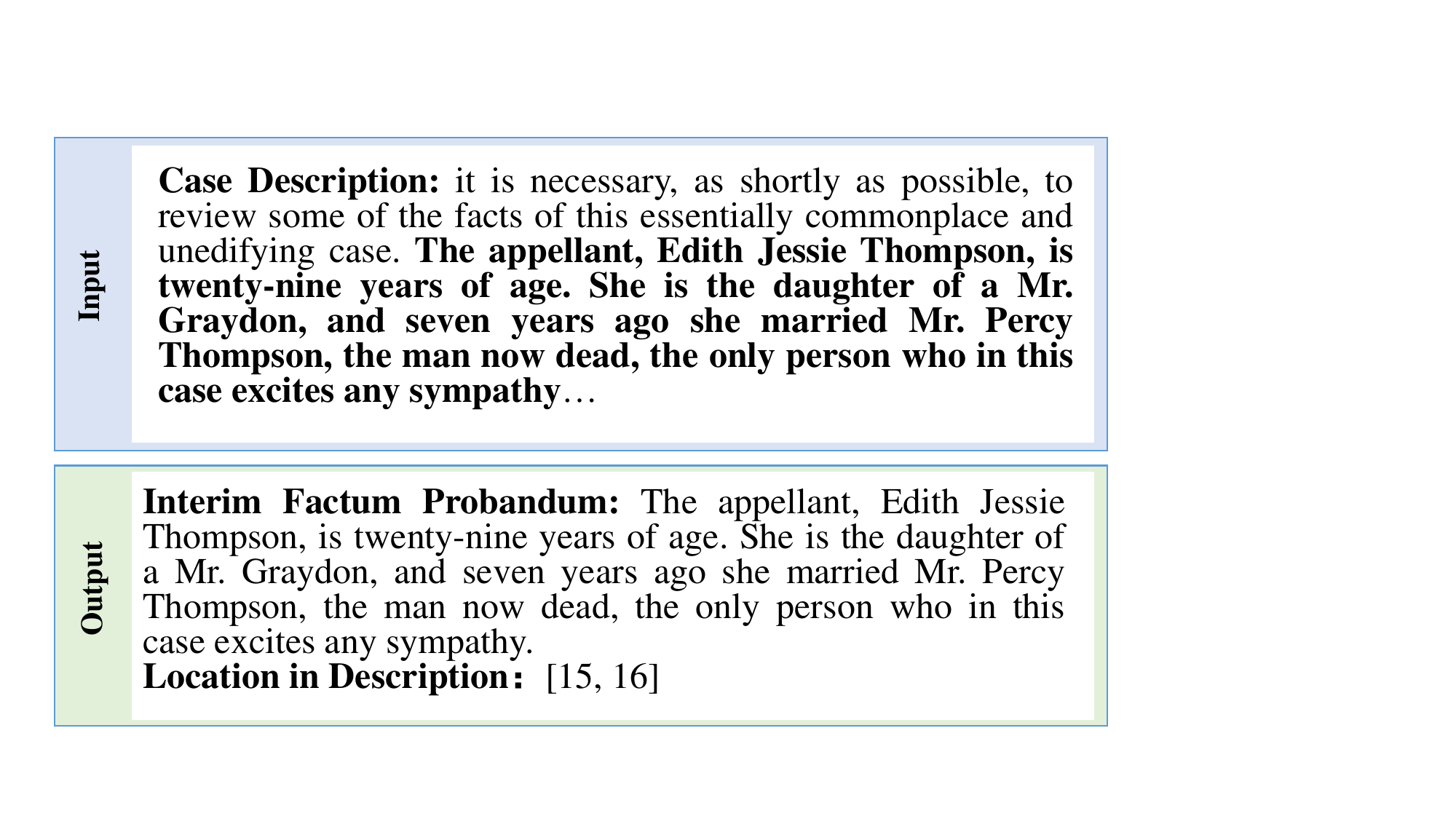}  
\caption{Illustration of the factum probandum generation.}
\label{fig:subtask1}
\vspace{-5mm}
\end{figure}

We propose our task, \textbf{T}ransparent \textbf{L}aw-Reasoning with Tree-Organized Structures (\task for short), which aims to generate the hierarchical and layered fact-finding structure from the unstructured textual case description, as shown in Figure \ref{fig:task}. The law reasoning structure should follow our designed schema, but we limit facts to only the two dimensions of Interim probandum and ultimate probandum due to the difficulty of identification and labeling.
Formally, we aim to find a model $\mathcal{M}$, which takes the textual case description\footnote{The case description is a brief account of the case, usually including the times, events, actions, or behavior of each party, and any other important details that are relevant to the case.} $\mathbf{x} = [x_1, ...,x_n]$ with $n$ tokens as input and predicts the law reasoning structure $\mathbf{y}$, i.e., $\mathbf{y} = \mathcal{M}(\mathbf{x})$. Note that, the ground-truth structure is labeled following the schema defined in Section \ref{sec:schema_formulation}.
In detail, \task includes four sub-tasks according to its three elements (i.e., factum probandum, evidences, experiences, and inferences). We introduce each sub-task as follows:

\paragraph{Sub-task I: Factum Probandum Generation}
Aim to generate the factum probandum $\textbf{F}$ that comes from a case description $\textbf{x}$, including interim probandum, and ultimate probandum. Among them interim probandum can be extracted from the case description and ultimate probandum should be generated in other ways. Figure \ref{fig:subtask1} shows an example to locate interim probandum in a case description.



\paragraph{Sub-task II: Evidence Reasoning}
Aim to specify the evidence that supports the interim probandum. For each interim probandum, multiple pieces of evidences are directly extracted from the case description. 

The sub-task aims to find a model $\mathcal{V}$, which takes the case description $\mathbf{x}$  and factum probandum query $\mathbf{q_f}$ as input to extract the corresponding $\mathrm{v}_i$, i.e., $\mathbf{v}_i = \mathcal{V}(\mathbf{x}, \mathbf{q_f})$. So this sub-task actually corresponds to the evidences and references elements in the schema.

This task can be divided into two sequential steps: the first step involves extracting evidence from the case description, and the second step entails linking the extracted evidence to the interim probandum.

\begin{figure}[h]
\centering
\includegraphics[width=1\linewidth]{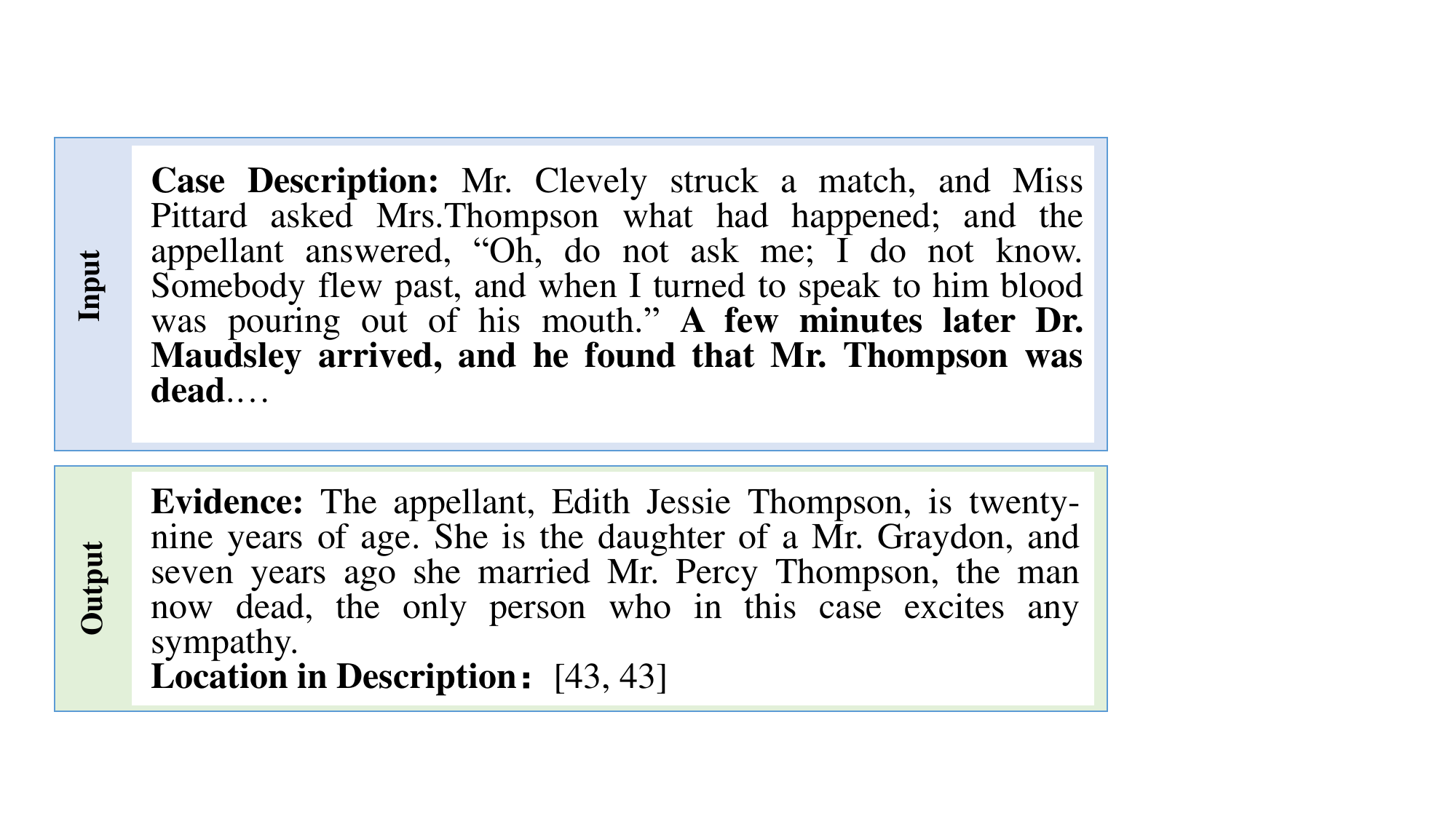}  
\caption{Illustration of the evidence extraction in subtask 2.}
\label{fig:subtask2_extract}
\vspace{-5mm}
\end{figure}

Figure \ref{fig:subtask2_extract} shows an example of step 1. Each evidence $\mathrm{v}_i$ is a span $[p_s, p_e]$, with $p_s, p_e$ indicating the beginning and ending position in the case description. The evidence is localized at the sentence level.

\begin{figure}[h]
\centering
\includegraphics[width=1\linewidth]{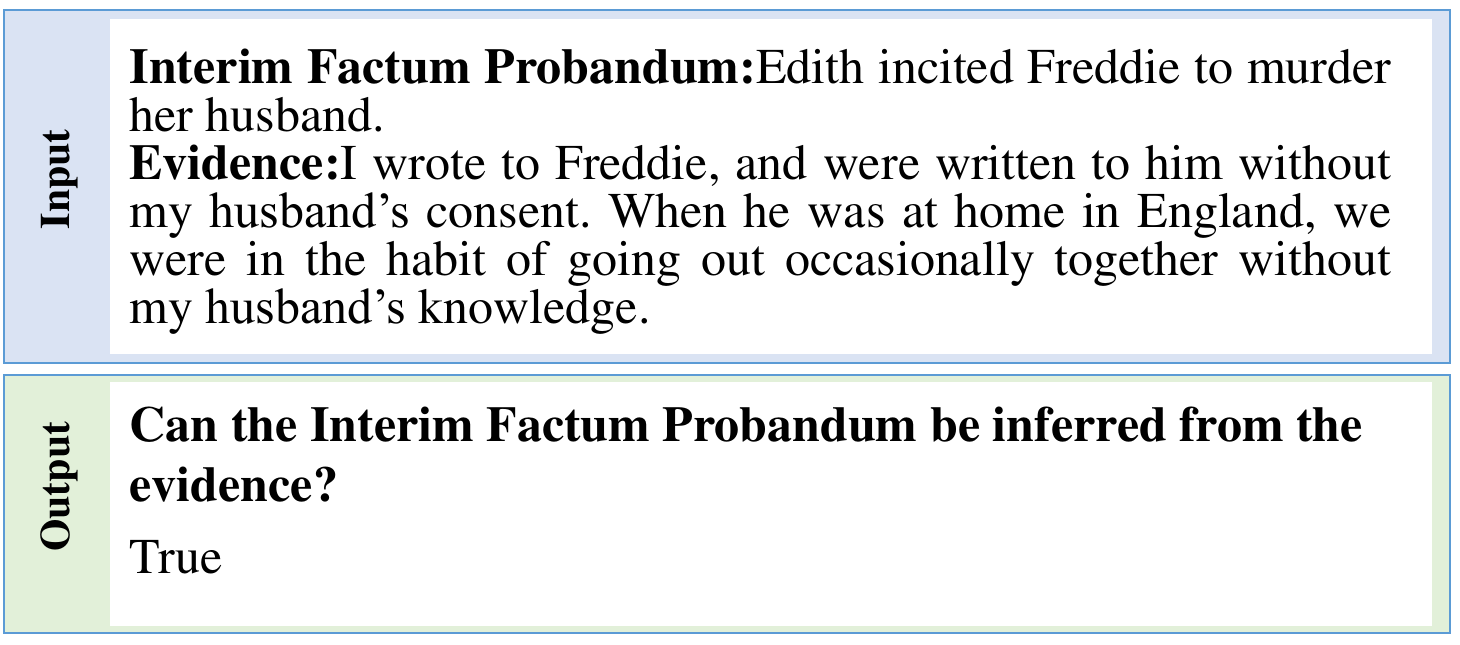}  
\caption{Illustration of the evidence reasoning in subtask2.}
\label{fig:subtask2_reason}
\vspace{-5mm}
\end{figure}

The process of step 2 is shown in Figure \ref{fig:subtask2_reason}. We contribute the relationship between the evidences extracted from the case description in the previous step and the interim probandum. If the interim probandum can be inferred from the evidence, we consider that a connection exists between the evidence and the interim probandum.

\begin{figure}[h]
\centering
\includegraphics[width=1\linewidth]{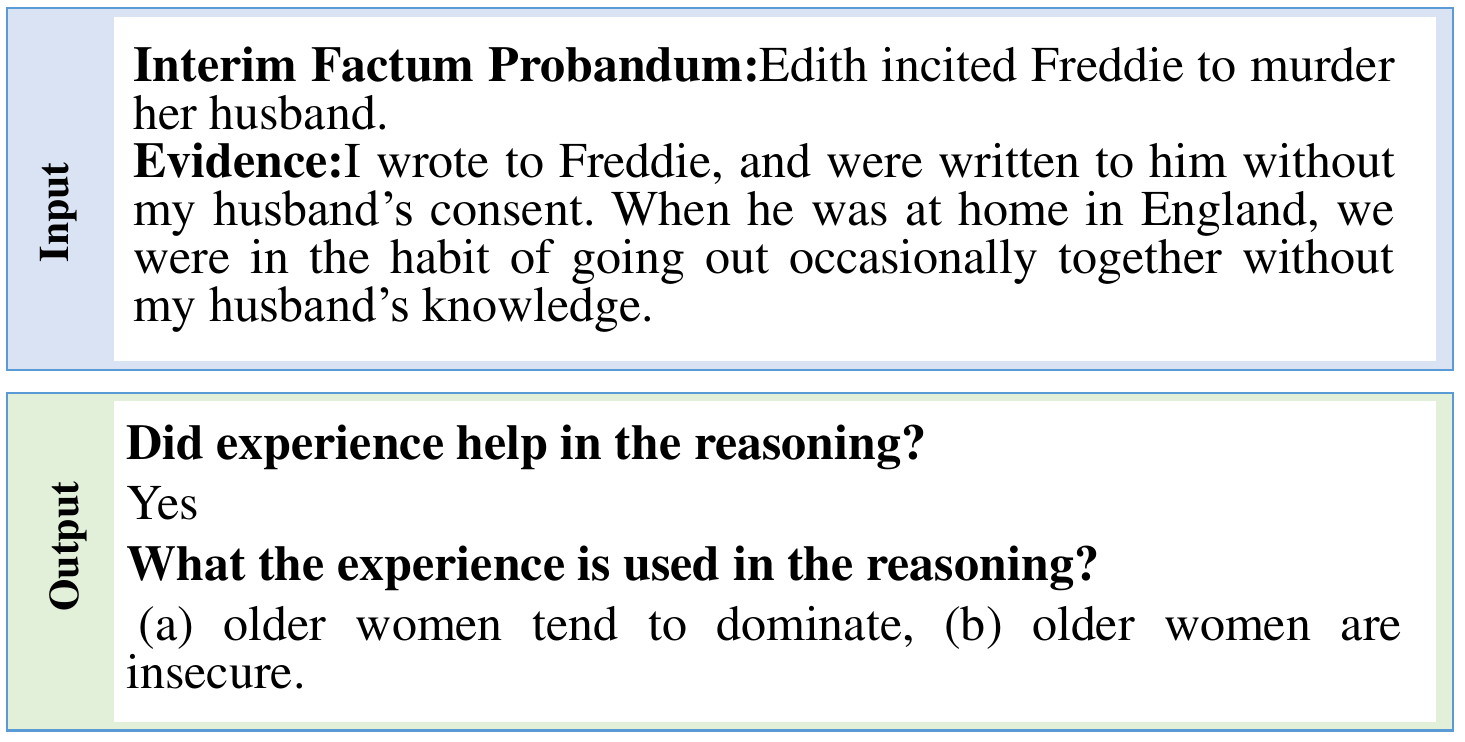}  
\caption{Illustration of the experience generation.}
\label{fig:subtask3}
\vspace{-5mm}
\end{figure}

\paragraph{Sub-task III: Experience Generation}
Aim to reveal the human experiences $\mathrm{e}$ between the evidences $\mathrm{v}$ and the interim probandum $\mathrm{f}$. Figure \ref{fig:subtask3} shows an example.






\section{Dataset Construction}\label{sec:data_collection}

We construct a high-quality dataset of case descriptions with multiple levels of annotated factum probandum, evidence and links with correlative facts, and the involved experiences, which follow our schema and show the explicit path of the law reasoning process. This section delves into the details of the crowd-sourcing construction, statistical analysis and quality control(in Section \ref{sec:quality_control}). We utilize publicly available data and implement strict controls over the annotation process. You can get more details on bias and ethical considerations in Section \ref{sec:dataset_statement} from the appendix.  

\subsection{Crowd-Sourcing Dataset}\label{sec:data_analysis}

We collect the unannotated data from China Judgement Online Website and each sample from the unannotated data describes a real-world case. Then we employ a two-phase methodology inspired by the Wizard-of-Oz technique \cite{kelley1984iterative,dahlback1993wizard} to gather the annotated dataset. In the initial phase, we carry out small-scale pilot studies in which we request multiple workers to annotate the documents and employ law professionals to assess the quality of their annotations. This helps us ensure the reliability of our labeling methods and summarizes a set of tips for labeling and testing quality. In the second phase, we train and certify a large group of crowd-workers through tutorials and qualification tests developed during the first phase. We pay them the local average hourly wage. The second phase includes two stages. 

\noindent\textbf{Stage I \textbf{Automatic Annotation} }We apply the automatic mechanism and follow the idea of the schema in Section \ref{sec:schema_formulation}. We use a prompt-based approach to label the unannotated data. The approach can be decomposed into three steps, each containing several turns of QA, which refer to the dialogue with Large Language Models (LLMs). By these steps, we find out the factum probandum, evidences, the facts they support,, and the corresponding experiences in turn. 

\noindent\textbf{Stage II \textbf{\textbf{Human-refinement} } }
We create a web-based survey using Label Studio\footnote{\url{https://labelstud.io}},a data labeling platform that allows workers to refine data generated from the automatic stage. We train workers with a comprehensive set of instructions and a well-defined example to collect high-quality rewrites that follow the schema and fulfill the requirements of judicial officers. During the refinement stage, we will present workers with the original description of a single case along with the corresponding factum probandum, evidences and links, and experiences obtained from Stage I. Workers are requested to relabel the data with the inference of labeled data annotated by LLMs. This helps accelerate the labeling speed and does not cause any false negatives.

\subsection{Dataset Statistical Analysis}\label{sec:data_analysis}
The collected data comprises 453 cases, 2,627 factun probandum, 14,578 pieces of evidence, and 16,414 experiences. The total number of tokens in the dataset is 6,234,443. The data statistics of the dataset are shown in Table~\ref{tab:data-analysis}. It is noteworthy that, we can construct an instruction dataset with a scale exceeding 40,000 samples, which can be utilized for fine-tuning LLMs.

\begin{table}[h!]
    \centering
    \resizebox{0.48\textwidth}{!}{%
        \begin{tabular}{l|c|c|c}
            \hline \toprule
                                 & \textit{Train} & \textit{Val} & \textit{Test} \\ \hline
            \# Instances         &  253        & 100        & 100         \\ \hline
            \# Tokens            &  3,877,780    & 897,916      & 1,458,747       \\
            \hline
            \# Ave. Evidences       & 36.05          & 15.30         & 39.28         \\ \hline
            \# Ave. Facts & 6.77           & 3.47         & 5.67          \\ 
            \hline
            \# Ave. Experiences & 37.77   & 18.63        &44.5
                \\
            \hline
            \# Ttl. Evidences      &
            9,120           & 1,530          & 3,928
                \\
            \hline
            \# Ttl. Facts & 1,713          & 347          & 567
                \\
            \hline
            \# Ttl. Experiences & 9,550
              & 1,863         & 4,450
                \\ \bottomrule
        \end{tabular}%
    }
    \caption{Data analysis of the collected data. Ave.: Average. Ttl.:Total. \#: The number of $\cdot$.}
    \label{tab:data-analysis}
\end{table}

\section{Approach}\label{sec:approach}
For the task, we propose our Knowledge-enhanced Transparent Law Reasoning Agent (\textit{\task Agent}). This approach, which see the whole law reasoning process as a tree structure, adheres to the established analytical approach employed by legal professionals. 

\begin{figure*}[htp]
\centering
\includegraphics[width=1\linewidth]{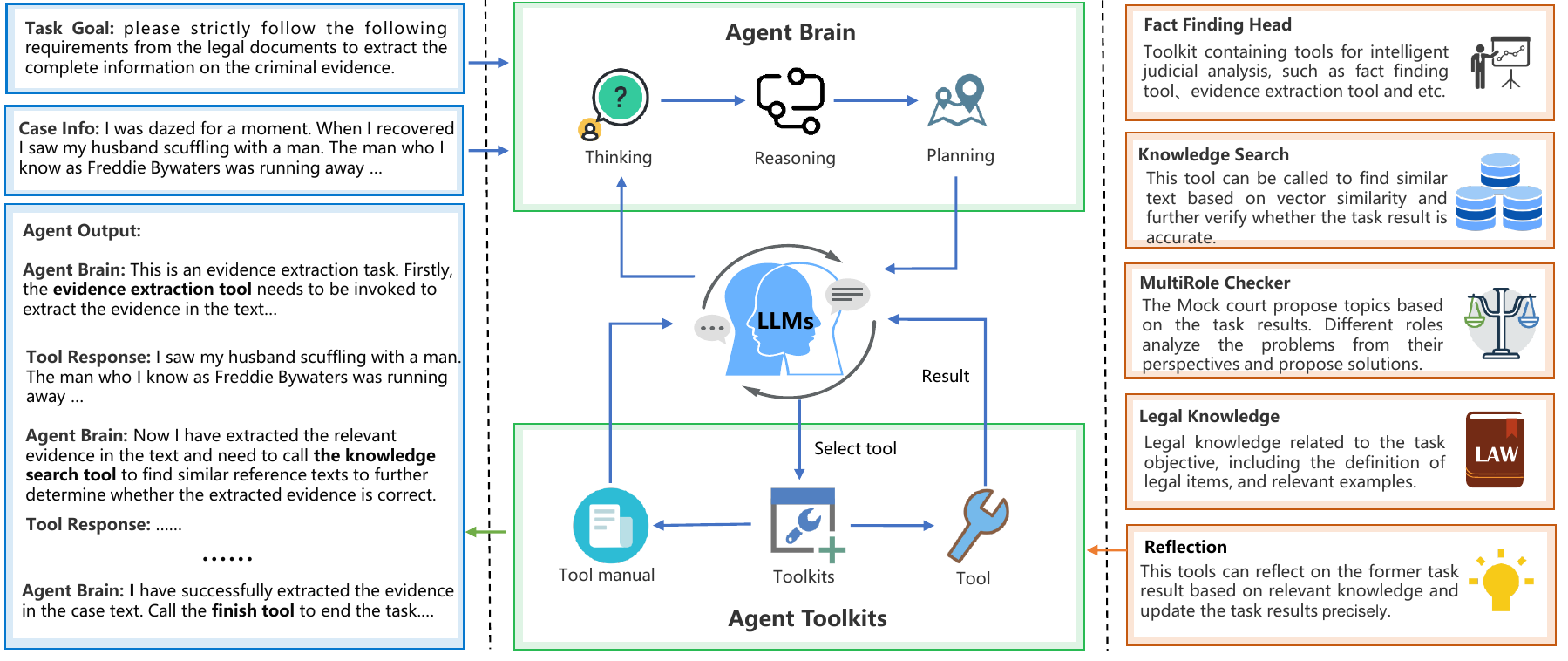}  
\caption{Illustration of our approach.}
\label{fig:framework_illustration}
\vspace{-5mm}
\end{figure*}

As illustrated in Figure \ref{fig:framework_illustration}, the left side of the diagram depicts the user's input and the agent's output, the middle section outlines the fundamental workflow of the agent, and the right side present the main toolkit of the agent. 

The task objective and corresponding case information are input into the agent, which, based on the tool manual's instructions, progresses through the stages of thinking, reasoning, and planning. Subsequently, the agent selects the appropriate tools from the toolkit to accomplish the task. To enhance the quality of the outcome, the agent analyzes the returned results to determine if additional tools are necessary to improve accuracy. This process continues until the agent deems the results satisfactory, at which point it invokes the finish tool to conclude the task.

In the following sections, we will delve into the details of our agent toolkit in Section \ref{agent_toolkits} and agent strategy in Section \ref{agent_strategy}. The prompt design details of each tool can be referred to in Section \ref{sec:toolkits_details} in the appendix.

\subsection{Designed Toolkits}\label{agent_toolkits}

\noindent \textbf{Fact Finding Head} The toolkit is designed with various tools for handling legal documents based on different task objectives in Section \ref{sec:task_formulation}, including fact extraction tool, evidence extraction tool, fact-evidence linking tool, experience generation tool, etc. These tools define the task objectives, task rules, and json output format in the prompt to ensure that the task results are output in parallel when the tools are called, to ensure the accuracy and efficiency of the output results, and at the same time to facilitate the subsequent analysis and processing of the results by the model or the user.
 
\noindent \textbf{Knowledge Search} This toolkit contains different vector databases, which can retrieve similar texts in the vector database based on the input query to assist in determining whether the text meets the task objective. The tools in this toolkit have two steps. The first step is to retrieve similar texts based on the query, and the second step is to input the similar texts and the query into the LLM to comprehensively determine whether the query conforms to the task objective.

For example, in the extraction task of factum probandum generation, we first use the extracted interim probandums as the query to retrieve similar texts in the vector database. Each similar text has a corresponding binary classification label. True indicates that this text belongs to the interim probandums, and False indicates that this text does not belong to the interim probandums. If there are more texts belonging to the interim probandums among the similar texts, the LLMs is more inclined to consider the input query as a interim probandums.
 
\noindent \textbf{MultiRole Checker} The agent will throw out issues based on the task objectives and task results in the previous step and provide them to this toolkit. The LLMs in this toolkit will respectively play different roles such as lawyers, judges, and police officers, analyze the issues from different perspectives, provide suggestions for solving the issues, and vote to determine whether the quality of the task completion is excellent.

\noindent \textbf{Legal Knowledge} Legal knowledge related to the task objective, including the definition of legal items, legal provisions and relevant examples.
 
\noindent \textbf{Reflection} This toolkit can reflect on whether the task result is accurate based on the task objective and the knowledge of other tools. It is mainly used as a middleware for the invocation of other tools, which can effectively ensure the consistency of the task output result and the accuracy of the task result.

There are some other tools used by our agent. Compared with the above-mentioned tools, these tools may be used less frequently.

\noindent \textbf{Emotion Check} This tool can determine the sentiment of the input text. There are three labels: positive, negative, and neutral. In factum probandum generation, it can be used to discriminate whether the generated facts contain sentiment to ensure the objectivity and neutrality of the generated facts.

\noindent \textbf{Pattern Match} This tool can automatically pre-analyze the required knowledge, rules, and text features that meet the task objective.

\noindent \textbf{Finish} When the agent judges that the generated result meets the task, this tool is called to finish the task.

\subsection{Agent Strategy}\label{agent_strategy}

To guide LLMs to leverage these powerful tools properly, we develop agent strategies to efficiently complete our task objective. Our agent's strategy is a ReAct-like~\citep{yao2022react} strategy. This strategy prompts large language models (LLMs) to generate reasoning traces and task-related actions in an interleaved manner. Depending on these actions, LLMs choose appropriate external tools and call them by supplying relevant inputs. Subsequently, this strategy regards the outputs of the tools as additional knowledge and determines whether to call a final tool or other tools for further processing.

Specifically, after each invocation of knowledge-based tools (excluding the reflection tool) 1 to 2 times, The LLMs will call the reflection tool to conduct reflection and update the task results by integrating the previously obtained knowledge. This approach can not only ensure that the results returned by the knowledge-based tools are fully utilized, thereby improving the accuracy of the task results, but also maintain the consistency of the format of the task results.

\begin{table*}[htp]
\centering
\resizebox{1.0\textwidth}{!}{%
\begin{tabular}{l|ccc|ccc|ccc|c}
\hline \toprule
\multirow{2}*{\textbf{Approach}} & \multicolumn{3}{c|}{\textbf{Task I}} & \multicolumn{3}{c|}{\textbf{Task II}} & \multicolumn{3}{c|}{\textbf{Task III}} & \textbf{All} \\ 
\cline{2-11}
& $S_{fact}$-\textit{1} & $S_{fact}$-\textit{2} & $S_{fact}$-\textit{l} &$Pre$ & $Rec$ & $F_{evi}$ & $S_{exp}$-\textit{1} & $S_{exp}$-\textit{2} & $S_{exp}$-\textit{l} & $S_{c}$ \\ 
\hline
ChatGLM-6B&18.26&6.70&15.8&3.65&7.42&4.89&20.69&4.76&12.2&11.54 \\
LexiLaw&18.65&7.59&15.98&2.51&12.97&4.20&16.60&3.58&13.80&12.56 \\
Lawyer Llama v2&21.52&9.60&18.89&1.45&5.80&2.23&11.55&2.18&5.40&10.56 \\
\midrule
ChatGLM-6B finetune&29.30&19.11&26.82&5.95&23.56&9.50&23.12&4.26&19.17&14.37
 \\
Lexilaw finetune&29.91&20.40&26.57&8.87&27.09&13.37&19.37&2.41&	16.69&23.46 \\
Qwen-6B finetune&30.6&21.3&27.54&8.02&11.21&9.34&11.21&9.34&13.45&20.52 \\ 
\midrule
Spark 4.0 Ultra&25.61&13.33&22.33&7.62&6.66&7.11&23.54&5.44&18.31&24.63 \\
ERNIE-4.0 Turbo-8k&26.83&13.16&22.37&5.26&7.66&6.24&28.7&8.53&22.31&26.38 \\
Qwen-max&25.01&12.60&21.53&12.28&15.90&13.85&27.84&6.83&21.25&30.94 \\
GLM-4-plus&23.23&10.33&19.70&9.65&18.96&12.78&25.75&5.61&20.60&26.43
 \\
Deepseek-v3&29.47&14.89&25.73&10.74&19.10&13.75&\textbf{31.61}&\textbf{9.21}&\textbf{25.53}&30.35
 \\
\midrule
Claude-3.5&28.69&14.47&25.43&2.94&4.79&3.64&19.89&1.82&15.54&23.92 \\
GPT-4o-mini&28.98&14.92&25.16&4.48&13.04&6.69&27.6&5.77&21.71&24.69 \\
GPT-4o&29.86&16.43&26.44&9.72&19.84&13.05&28.71&7.31&22.36&25.74 \\
\midrule
\midrule
\task Agent&\textbf{32.99}&\textbf{18.03}&\textbf{28.75}&\textbf{10.38}&\textbf{40.73}&\textbf{16.53}
&30.92&8.66&24.81&\textbf{31.50} \\
\bottomrule
\end{tabular}%
}
\caption{Comparison between our approach and baseline models. We use the comprehensive score to assess the whole structure. $S_{c}$: Comprehensive score. We also list the performance of three sub-tasks. The numbers \textit{-1}, \textit{-2}, and \textit{-l} after \textit{S} correspond to Rouge-1, Rouge-2, and Rouge-l in the formula respectively.}
\label{tab:totaltaskresult}
\vspace{-3mm}
\end{table*}

\section{Experiments}\label{sec:exp}
Our agent has been rigorously compared against several classic LLMs available on the market, as detailed in Section \ref{sec:results}. Additionally, we conducted a comprehensive comparison with state-of-the-art reasoning models, including o1 and r1, as discussed in Section \ref{sec:reasonexp}. To further validate the effectiveness of our tools, we performed an ablation study, the results of which are presented in Section \ref{sec:ablation_study}. These comparisons and analyses collectively demonstrate the robustness and superior performance of our agent in various tasks and scenarios.

\subsection{Setup}\label{sec.setup}
\paragraph{Dataset}The test set in our experiments uses the dataset we constructed in Section \ref{sec:data_analysis}. The training set used when fine-tuning the model is the training set and validation set we constructed in Section \ref{sec:data_analysis}. We split each case content into multiple fragments, with the length of the content being no more than 1500 tokens, and then constructed an instruction dataset including 5w samples with the corresponding evidence, factum probandum and experiences. The constructed dataset is used to finetune LLM.

\paragraph{Metrics} 
For task 1, task 3, and the comprehensive evaluation (All), the results are assessed using a modified version of the rouge score\. For task 2, the evaluation is conducted based on precision, recall, and f1 metrics. The definitions of these metrics are provided in Section \ref{sec:metrics}. 

\paragraph{Agent Setting}
The base model used by the agent is 4o-mini\footnote{https://platform.openai.com/docs/models/gpt-4o-mini-2024-07-18}. The basic parameters of the model are temperature of 0.6, max\_tokens of 8096, top\_p of 1, frequency\_penalty of 0, presence\_penalty of 0, and number\_of\_results of 1. The vector database used by the Agent is chroma\footnote{https://www.trychroma.com/}, the vector model is bge-large-zh-v1.5\footnote{https://huggingface.co/BAAI/bge-large-zh-v1.5}, and the database is Postgre SQL 16\footnote{https://www.postgresql.org/}.

\paragraph{Baselines}\label{sec:baseline}
We compare our approach with strong baselines: 

The first group of baseline models comprises models with fewer than 13B parameters, which have not undergone fine-tuning using the task-specific \task dataset. Notably, the \textbf{ChatGLM-6B}~\citep{du2022glm} model has not been fine-tuned on the legal domain dataset. The \textbf{Lexilaw}~\citep{LexiLaw} model, however, is a variant of ChatGLM-6B that has been fine-tuned with legal domain-specific data. Similarly, the \textbf{Lawyer Llama v2}~\cite{huang2023lawyer} model is an outstanding open-source Chinese legal LLMs and the model is a fine-tuned version of the Llama3 model, adapted to the legal dataset.

The second group encompasses models that have been fine-tuned using the \task dataset. Specifically, instruction datasets, as described in Section \ref{sec.setup}, were used to fine-tune the ChatGLM-6B, Lexilaw, and Qwen~\citep{qwen2.5} models, producing their respective fine-tuned variants.

The third group consists primarily of API-accessible LLMs, which have been trained predominantly on Chinese language corpora. This group includes models such as \textbf{Spark 4.0 Ultra}\footnote{https://xinghuo.xfyun.cn/}, \textbf{ERNIE-4.0 Turbo-8k}, \textbf{Qwen-max}\footnote{https://qwenlm.github.io/blog/qwen2.5-max/}, \textbf{GLM-4-plus}\footnote{https://bigmodel.cn/}, and \textbf{Deepseek-v3}~\citep{deepseekai2024deepseekv3technicalreport}.

The fourth group features API-based LLMs trained primarily on English corpora, including \textbf{Claude-3.5}, \textbf{GPT-4o-mini}, and \textbf{GPT-4o}~\footnote{https://platform.openai.com/docs/models/gpt-4o-2024-08-06}.

To ensure optimal performance for various tasks, specific prompts have been designed to guide these API-based LLMs in completing \task tasks efficiently.

\subsection{Results}\label{sec:results}

\noindent\textbf{Comprehensive Score.}
It can be observed from Table \ref{tab:totaltaskresult} that \task agent not only addresses the issue of producing the irrelevant experiences but also enhances the precision of the extracted evidence and generated factum probandum by incorporating supplementary legal knowledge and legal processes.

\noindent\textbf{Factum Probandum Generation.} 
As shown in Table \ref{tab:totaltaskresult}, our agent model, through multi-step reasoning and tool utilization, effectively extracts interim probandum from text and generates ultimate probandum. By employing our agent, even with the base model being 4o-mini, we achieve performance surpassing that of the gpt-4o model. Notably, a smaller model with 6B parameters, after fine-tuning on the \task dataset, demonstrates capabilities comparable to, or even exceeding, those of LLMs. Additionally, we observe that among the 6B parameter models not fine-tuned on the TL dataset, those fine-tuned with legal knowledge, such as Lexilaw and Lawyer Llama, outperform the ChatGLM model, which lacks such legal knowledge fine-tuning.

\noindent\textbf{Evidence Extraction.} 
\task agent demonstrates enhanced precision in extracting evidentiary statements from text and establishing accurate correlations between evidence and interim probandum. Furthermore, the results indicate that all models face significant challenges in tasks involving linkage identification between evidence and interim probandum. Our agent and baseline models exhibit a propensity to associate evidence with interim probandum redundantly, irrespective of the existence of a substantive inference relations, which consequently results in lower precision metrics across all models. This phenomenon underscores the inherent complexity and challenge of the task at hand.

\noindent\textbf{Experience Generation.}
Our agent model is capable of generating precise human-experience-based information necessary for inferring interim probandum from evidence, achieving performance comparable to that of DeepSeek-V3.

From the experimental results, it is observed that although some models (such as ChatGLM-6B fine-tune) have been fine-tuned for Task 3, their performance on Task 3 still does not surpass that of LLMs accessed via APIs (such as Deepseek-V3, GPT-4o-mini, and GPT-4o). This suggests that Task 3 relies on extensive commonsense knowledge and social experience, which are inherently embedded in larger-scale LLMs. 


\section{Conclusion}\label{sec:conclusion}
Artificial Intelligence legal systems currently face challenges in law reasoning. To address this issue, we propose \task agent for law reasoning. By following two key steps: schema design and establishing tree-organized structures process (i.e., evidential reasoning), we can develop an abstract, systematic, and formalized reasoning process for law reasoning tree based on unstructured data. This law reasoning system can serve as a foundation for the advancement of AI legal systems, enabling them to make judgments transparent.

To ensure transparency in the judge's decision-making process, it is important to visualize the experience relied upon and the intermediate conclusions reached at each step of reasoning and judgment. This serves as a helpful reminder to the judge of which experience was utilized in each step, thereby mitigating the inherent risk of personal bias and enhancing the accuracy of law reasoning along with the final judgment. Our contribution in terms of task formulation, dataset and modeling pave the way for transparent and accountable AI-assisted law reasoning.



\section*{Limitations}\label{sec:limitation}
Although \task agent has yielded impressive results, the underlying reasons for these outcomes have not been thoroughly investigated. Moreover, the use of open-ended natural language as prompts presents both advantages and challenges. Successful extraction often necessitates domain expertise to design schema and can be a time-intensive process.

\section*{Ethics Statement}\label{sec:ethics}
This study strictly adheres to the ethical principles outlined in the Declaration of Helsinki, which serves as a guiding framework for conducting research involving human subjects. It is of utmost importance to ensure that all participants in this study are treated with respect, dignity, and fairness.

To ensure transparency and informed decision-making, all participants will receive comprehensive information regarding the nature and purpose of the study. They will have the opportunity to ask questions and clarify any concerns they may have before providing their written informed consent. It is essential to emphasize that participation in this study is completely voluntary, and individuals have the right to withdraw their involvement at any point in time without facing any negative consequences or penalties.

In compliance with applicable laws and regulations, the confidentiality and privacy of all participants will be diligently protected. Measures will be implemented to safeguard their personal information and ensure that only authorized personnel have access to it. Any data collected throughout the study will be anonymized, ensuring that the identities of participants remain confidential.

By upholding these ethical principles and safeguards, we aim to conduct a study that upholds the highest standards of integrity and respects the rights and well-being of every participant involved.
\bibliography{acl2024}
\bibliographystyle{acl_natbib}

\appendix

\begin{table*}[ht]
\newcolumntype{C}{>{\ttfamily}l}
\newcolumntype{Y}{>{\RaggedRight}p{13cm}} 
\centering
\begin{tabular}{@{} C Y @{}} 
\toprule
\multicolumn{1}{c}{\textbf{Term}} & 
\multicolumn{1}{c}{\textbf{Definition}} \\
\midrule

Factum Probandum     &  The fact that must be proven. It's used in legal contexts to refer to a fact or set of facts that one party in a case must establish in order to prove their claim or defense. \\
        
Interim Probandum    & The provisional or temporary facts to be proven. It refers to facts that are temporarily or provisionally considered to be established for the purposes of an ongoing legal proceeding, pending further evidence or a final ruling. \\
        
Ultimate Probandum    &  The "ultimate fact" or the final fact that must be proven in a case. It is the core fact or facts that are central to the resolution of the legal issue at hand. The ultimate probandum is the fact that, if proven, will ultimately decide the outcome of the case. \\
        
Criminal Evidence  & The information, objects, or testimony presented in a court of law to prove or disprove the factum probandum. \\
        
Human Experience & The understanding of human behavior, societal norms, and practical reasoning to resolve disputes and administer justice. It play significant roles in evaluating evidence, determination of factum probandum and making judicial decisions.  \\
\bottomrule
\end{tabular}
\caption{Legal Terms}
\label{tab:glossary}
\end{table*}

\section{Metrics}\label{sec:metrics}


\noindent \textbf{Metrics for Factum Probandum Generation.} 
We use the Rouge $F_1$ score, which is commonly used for text summarization tasks.
For each case, the fact set provided as ground truth is represented as $\mathrm{F}^* = [\mathrm{f}_1^*, \mathrm{f}_2^*,..., \mathrm{f}_n^*]$, while the prediction set generated by a model is denoted as $\mathrm{F} = [\mathrm{f}_1, \mathrm{f}_2,..., \mathrm{f}_m]$. The metric can be defined using the following formula:

$$
S_{fact} = \frac{1}{n} \sum_{i=1}^{n} \max_{\mathrm{f}_j^* \in F} (Rouge(\mathrm{f}_i, \mathrm{f}_j^*))
$$

\noindent \textbf{Metrics for Evidence Reasoning.}
The $F_{evi}$ metric measures how well a model extracts relevant evidences to support its factum probandum. It does this by comparing the model's predicted evidence spans to the actual ground-truth evidence spans and penalizing for both missing important evidence and including irrelevant information. Thus, each piece of evidence can be linked to specific factum probanda it supports. These connections are represented by triples (factum probandum, relation, evidence). Think of an arrow pointing from the evidence to the factum probandum. The model predicts some evidence to support the ground-truth facta probanda, resulting in triples. $F_{evi}$ focuses on the macro $F_1$-like metrics, meaning it only cares about how accurate the model's chosen triple is. The more overlap between predicted and ground-truth triples, the higher the score.
$F_{evi}$ is formulated as:
$$
Pre = \frac{\sum_{i=1}^{k} |\mathrm{L}^{*}_i \cap \mathrm{L}_i |}{\sum_{i=1}^{k}|\mathrm{L}_i|} 
$$

$$
Rec = \frac{\sum_{i=1}^{k} | \mathrm{L}^{*}_i \cap \mathrm{L}_i |}{\sum_{i=1}^{k}|\mathrm{L}^{*}_i|} 
$$

$$
F_{evi} =  \frac{2\cdot Pre \cdot Rec}{Pre+ Rec} 
$$
For each case, the triple set provided as ground truth is represented as $\mathrm{L}^{*}_i$. The prediction set generated by a model is denoted as $\mathrm{L}_i$. We use set intersection ($\cap$) to identify the overlap between the predicted and ground-truth set. $\alpha$ is a hyper-parameter to balance between $Pre$ and $Rec$. $k$ is the number of cases. $|\,|$ returns the number of the element in the set.

\noindent \textbf{Metrics for Experience Generation.}
The metric for experience generation considers two aspects. First, we should consider whether the experience needs to be generated as a component to achieve the interim probandum. It is a binary classification problem and we measure accuracy as the metric. Then, we consider the quality of the generated experience using Rouge $F_1$. The experience alone does not support an interim probandum. The following formula defines the process.

$$
R_{exp}(\mathrm{e}_i^*, \mathrm{e}_i)=\left\{
\begin{matrix}  
1 & \mathrm{e}_i^*= \mathrm{e}_i=None \\  
Rouge(\mathrm{e}_i^*, \mathrm{e}_i) &else 
\end{matrix}\right. 
$$

$$
S_{exp} = \frac{1}{t} \sum_{i=1}^{t} R_{exp}(\mathrm{e}_i^*, \mathrm{e}_i)
$$
$t$ is the number of generated experiences. $\mathrm{e}_i^*$ is the ground-truth experience quadruple (fact, relation, evidence, experience). $\mathrm{e}_i$ is predicted experience quadruple. $\mathrm{e}_i=None$ means that the relation from the evidence to the interim probandum doesn't require additional experience. If either $\mathrm{e}_i^*$ or $\mathrm{e}_i$ is not equal to $None$, $R_{exp}$ is set to 0.




\noindent \textbf{Comprehensive Score.} 
The three metrics mentioned above pertain to the sub-task level. To evaluate the comprehensive score, it is important to consider the overall quality of the structure, in addition to the necessity of each sub-task. The Comprehensive Score ($S_{c}$) is calculated as follows:
\begin{equation*}
 \begin{aligned}
Rouge_{sum} =  \frac{1}{2}(&Rouge(\mathrm{d}_m, \mathrm{d}_m^*) +  \\
& Rouge(\mathrm{d}_n, \mathrm{d}_n^*)) \\
\end{aligned}   
\end{equation*}

$$
\hat{r}_q = arg\max_{\mathrm{r}_q^* \in \mathrm{y}_i^*} (Rouge_{sum}(\mathrm{r}_p, \mathrm{r}_q^*))
$$

\begin{equation*}
 \begin{aligned}
S = \frac{1}{max([\mathrm{y}_i], [ \mathrm{y}_i^*])}\sum_{p=1}^{[ \mathrm{y}_i]} &(Rouge_{sum}(\mathrm{r}_p, \mathrm{\hat{r}}_q) + \\
&R_{exp}(\mathrm{e}_p, \mathrm{\hat{e}}_q))
\end{aligned}   
\end{equation*}

$$
S_{c} = \frac{1}{k}\sum_{i=1}^{k}S
$$
$\mathrm{y}_i$ is the predicted fact-finding structure. $\mathrm{y}_i^*$ is ground-truth structure. Each $\mathrm{y}_i$ include two basic elements, nodes $\mathrm{d}$ ($\mathrm{d} \in {\left \{ \mathrm{f}, \mathrm{v} \right \} }$) and relation $\mathrm{r}$, which connect between node $\mathrm{d}_m$ and node $\mathrm{d}_n$.
$[ ]$ denotes the number of relations in structure $\mathrm{y}$.

\section{Quality control}\label{sec:quality_control}

Since labeling is a task without formulaic standards, we employ multiple methods to control the annotation quality.

\noindent\textbf{\textbf{Data Source} }We use data from China Judgement Online Website\footnote{\url{https://wenshu.court.gov.cn}}, , which assures our case descriptions are following a relatively fixed logical structure. This reduces the difficulty of labeling, even that amateurs can also understand the idea of the annotation after receiving a little training. 

\noindent\textbf{\textbf{Workers and Payment}} We restrict the workers to those in law schools. Their research direction is highly aligned with the topic of our paper. 

In particular, we recruited a total of 45 students, with an hourly labor compensation of \$ 7.5. The average labeling time for each annotation is 55 minutes, and the verification time is 20 minutes. On average, each annotator was assigned to 15.1 annotations, and the reviewer was assigned to 30.2 annotations. The total labeling duration is 566.25 hours, and the total labor cost paid is \$ 4246.9.

\noindent\textbf{\textbf{Training and Pre-labeling}} 

We initially provided training to the recruited annotators, standardizing the annotation criteria. This ensured the accuracy and standardization of task element annotation. The training covered fundamental principles and formats for transparent law reasoning tasks, usage methods of the annotation system, and detailed issues related to annotations. To further improve the precision of annotation, we initially selected five cases for pre-labeling and offered guidance on addressing problems that arose during pre-labeling as well as bias issues observed in some annotation results.

\noindent\textbf{\textbf{Annotation Process}}
On average, there are 3 workers assigned to each annotation. During the labeling process, two annotators are responsible for labeling, while the third skilled worker verifies the evaluation results. In cases where there are disagreements among the two labeling workers, we collect the data and mark the disagreements. Any samples that have conflicting results are reviewed manually by another law professional. We also control the quality of workers by excluding the results made from workers whose work constantly appears to be inadequate. Other quality control measures are implemented too. Prior to survey submission, the refined data will be examined by the Label Studio to verify the edits made by the workers. If incomplete sentences are identified, a message will be displayed, prompting workers to review their work. 

In these ways, our labeled results show great reliability. After careful review, the inter-annotator agreement scores (specifically, inter-rater Spearman and Pearson Correlations) are found to be above 0.93, indicating a strong consensus.

\section{Bias and Ethics Statement}\label{sec:dataset_statement}

We use data sources that have been published on the official website. Although the case descriptions involve names, they do not contain key privacy information such as user contact information and identity ID. Note that if there is private information in the cases data, the parties can apply to the court to prevent its disclosure. Therefore, our dataset, based on publicly available data, does not involve an invasion of personal privacy.

All the participation in this study is completely voluntary, and the confidentiality and privacy of all participants will be diligently protected. In addition, in the process of manual labeling, we show the full content of the case description to the manual annotator, in order to prevent potential biases resulted from the LLM automatic labeling. We used legal professionals to conduct a test label and legal experts to assess whether bias was included. All annotators are trained to consciously avoid bias and unfairness.

We aim to use our methods to assist in case adjudication and to support the public. While we explore a clear process for determining facts, this does not imply that we encourage the use of LLMs to make final case decisions.

\section{The Toolkits Detail}\label{sec:toolkits_details}

\subsection{Thought Process}

\begin{figure*}[t!]
    \centering
    \includegraphics[width=0.8\linewidth]{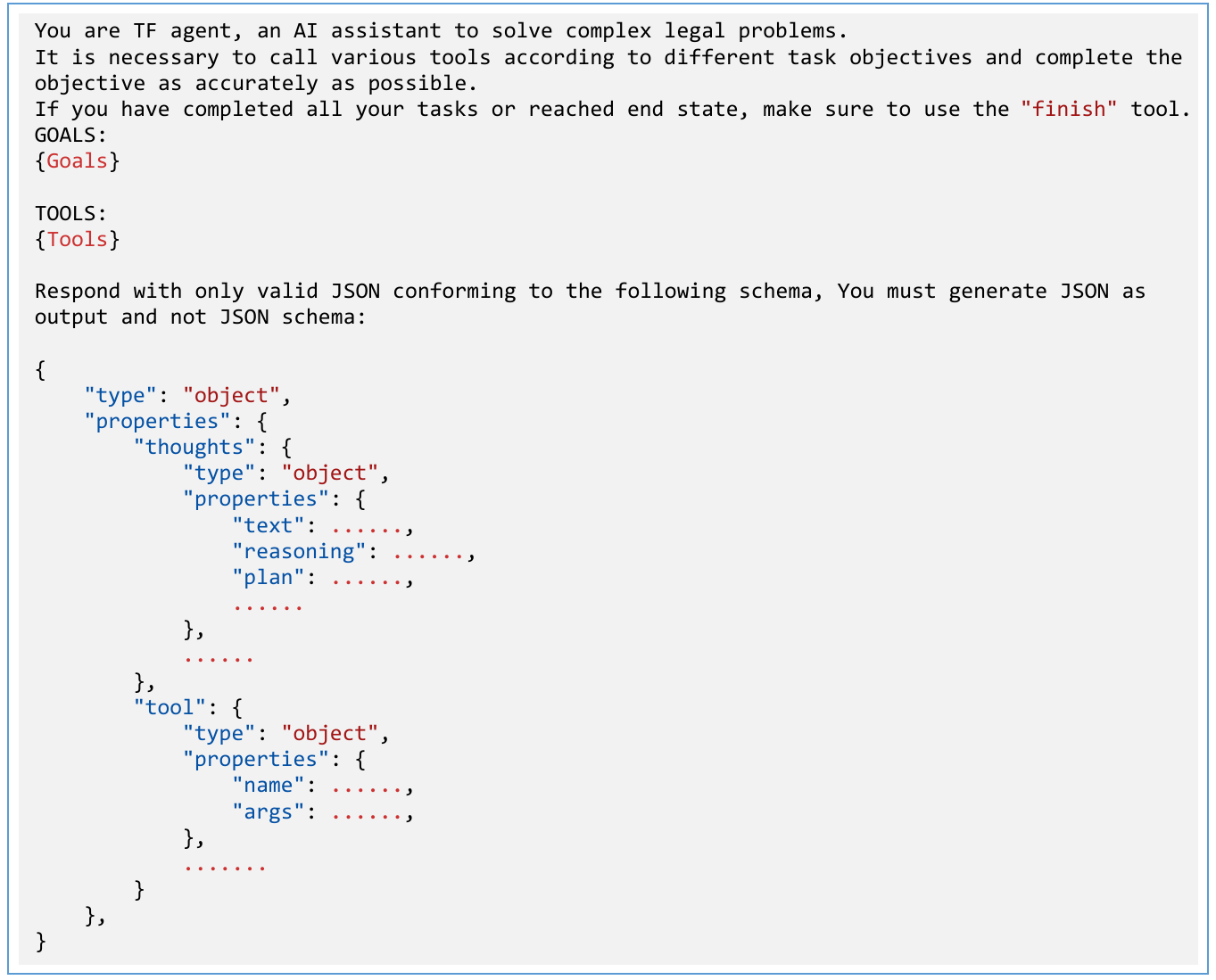}
    \caption{The thought prompt of \task agent .}
    \label{fig:prompt_thought}
    \vspace{-5mm}
\end{figure*}

The thought process and tool selection of the Agent are primarily controlled by LLMs, with the corresponding prompt illustrated in Figure \ref{fig:prompt_thought}. The first line clearly defines the problem the Agentis solving, the second line outlines the tool selection strategy, and the third line determines the termination conditions for \task Agent. Following this, the \texttt{\{Goals\}} field is used to input the objectives of the task along with relevant textual content. The \texttt{\{Tools\}} field enumerates the tools available for selection by the large model. The descriptions of these tools are generated by converting tool classes into textual representations using the Pydantic module, which includes the tool class name, class function, and the required input arguments.

Finally, the output format of the model is defined as a JSON-compliant string that can be successfully parsed by Pydantic. The returned JSON string include a "thinking" field for the model's reflections, which encompasses the thought content (text), reasoning (reasoning), and planning (plan) among other fields. The "tool" field specifies the name of the tool (name) to be invoked at the current step and the parameters to be passed (args).

This structured approach ensures a systematic and efficient decision-making process within the realms of artificial intelligence and deep learning, facilitating advanced computational tasks and analyses.

\subsection{Fact Finding Head}

\begin{figure*}[t!]
    \centering
    \includegraphics[width=0.8\linewidth]{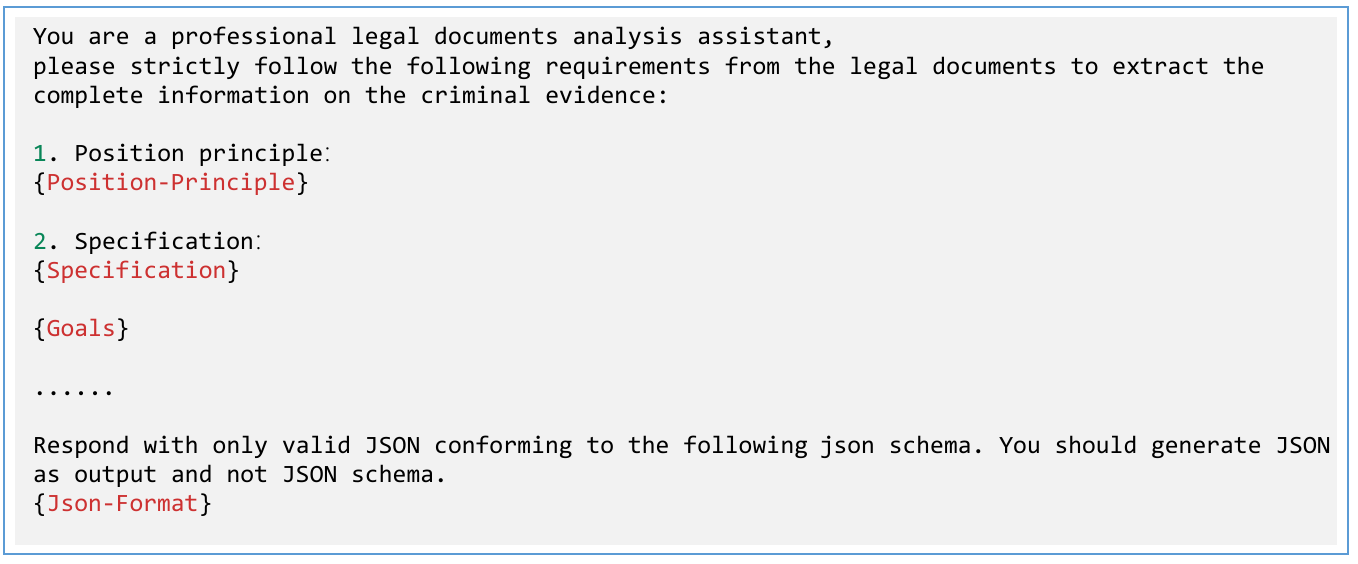}
    \caption{The evidence extraction prompt of \task agent .}
    \label{fig:prompt_evid_extract}
    \vspace{-5mm}
\end{figure*}

The Fact Finding Toolkits comprise five distinct tools, each specifically designed for different TL subtasks. These tools are capable of generating results in parallel and formatting the outputs, thereby enhancing the operational efficiency of the Agent and improving the quality of task results. Moreover, they facilitate the effective utilization of the Agent's results in experimental testing scenarios.

The first two tools are utilized in Task 1. The Interim Probandum Finding Tool generates an Interim Probandum based on the content of legal documents, while the Ultimate Probandum Generation Tool produces the final Ultimate Probandum from the obtained Interim Probandum. Tools three and four are applied in Task 2; the Evidence Extraction Tool extracts criminal evidence from legal documents, and the Evidence Linking Tool connects the criminal evidence related to the Interim Probandum. The Experience Generation Tool is employed in Task 3, which generates human experience from evidence reasoning to Interim Probandum.

These tools are driven by LLMs, and the corresponding prompts are illustrated in Figure \ref{fig:prompt_evid_extract}. Taking the Evidence Linking Tool as an example, the function of tool is clearly stated at the beginning. The \texttt{\{Position Principle\}} field informs the model of the potential positional and textual characteristics of the evidence, and the \texttt{\{Specification\}} field provides detailed specifications for the model's output. Finally, the model outputs a string that conforms to the \texttt{\{Json-Format\}} based on the objectives  \texttt{\{Goals\}}.

This structured approach ensures that the outputs are precise and tailored for further analysis and application in the fields of law.

\subsection{Multi-role Checker}

\begin{figure*}[t!]
    \centering
    \includegraphics[width=0.8\linewidth]{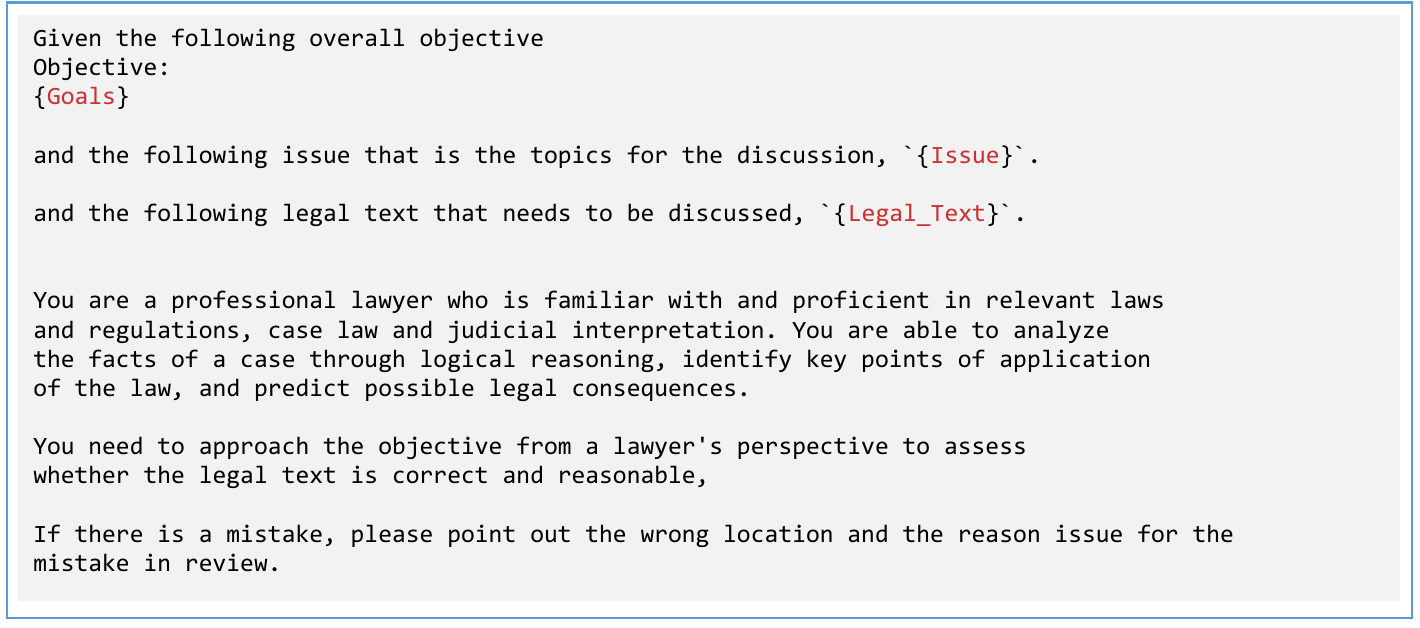}
    \caption{The laywer prompt of \task agent .}
    \label{fig:prompt_laywer}
    \vspace{-5mm}
\end{figure*}

\begin{figure*}[t!]
    \centering
    \includegraphics[width=0.8\linewidth]{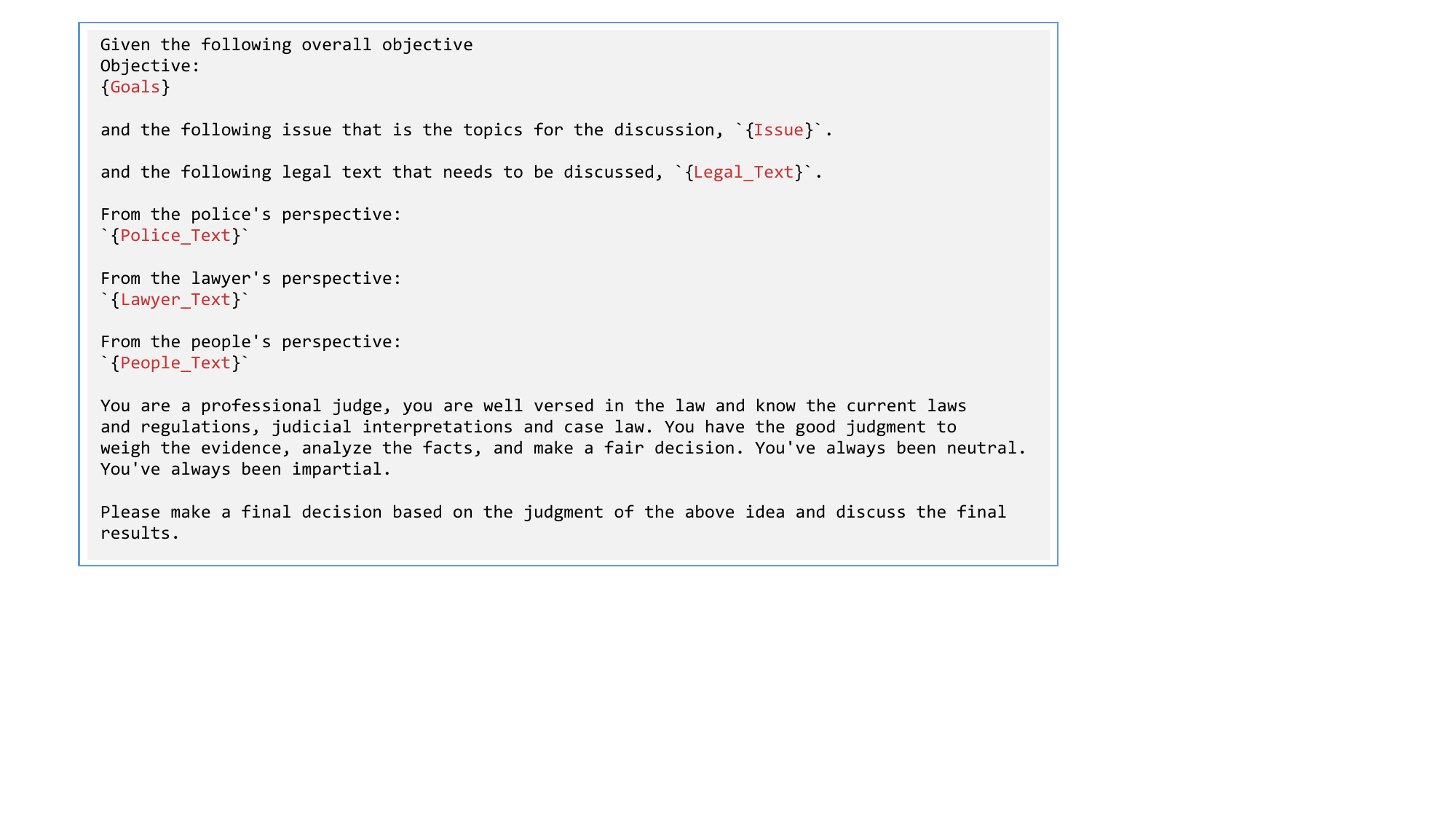}
    \caption{The judge decision prompt of \task agent .}
    \label{fig:prompt_judge_decision}
    \vspace{-5mm}
\end{figure*}

The Multi-role Checker is designed to address issues raised by the Agent by providing solutions through analyses comment from different roles . These analyses are synthesized based on the task objectives and the results generated by the \task Agent in previous steps. The Multi-role Checker tool operates in two main phases: the first phase involves different roles analyzing the problem and proposing solutions, while the second phase involves a chief justice synthesizing these solutions to arrive at a final decision.

In the first phase, we have defined three distinct roles—lawyer, police office, and general public—to analyze the problem. The basic prompt format for this phase is illustrated in Figure \ref{fig:prompt_laywer}. Here, the \texttt{\{Issue\}} field represents the question posed by the LLMs after deliberation based on previously generated results, and the \texttt{\{Legal\_text\}} field contains the text content under discussion. Subsequent paragraph describe the characteristics associated with each role. The last paragraph details the requirements expected from each role.

Finally, as illustrated in Figure \ref{fig:prompt_judge_decision}, we employ a prompt to consolidate the solutions proposed by the various roles. The judge role then synthesizes these inputs to deliver the final decision. This outcome is subsequently utilized to inform the Agent's subsequent thinking processes and tool selections, ensuring a coherent and well-considered approach to task execution.

\subsection{Reflection}

\begin{figure*}[t!]
    \centering
    \includegraphics[width=0.8\linewidth]{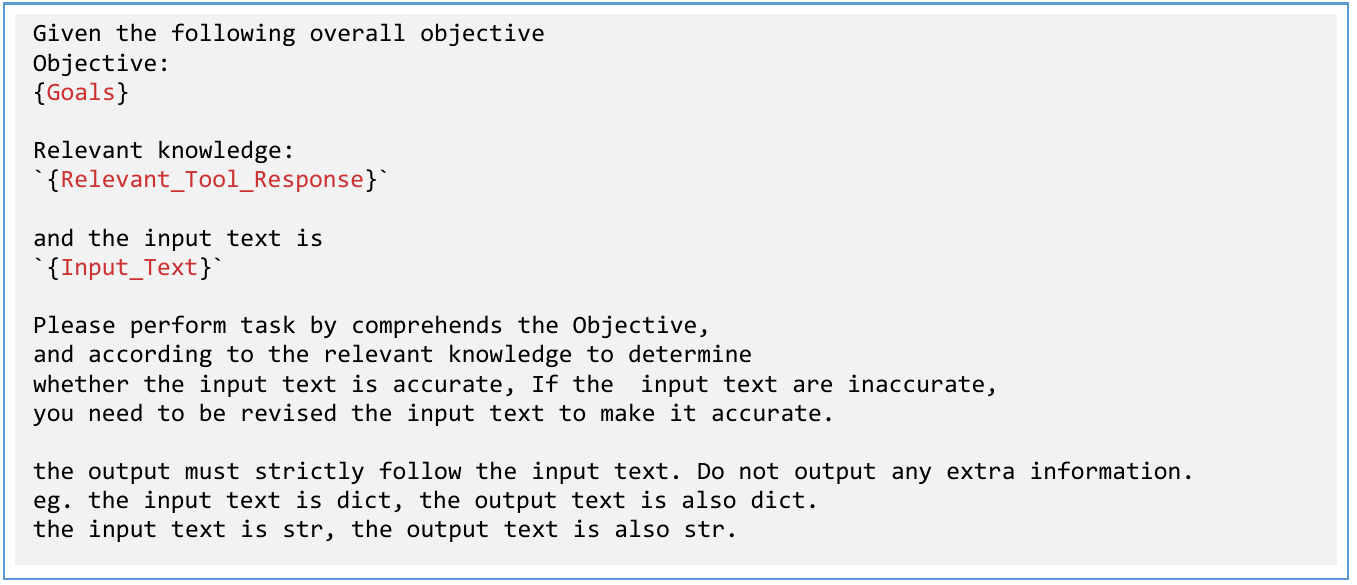}
    \caption{The reflection prompt of \task agent .}
    \label{fig:prompt_reflection}
    \vspace{-5mm}
\end{figure*}

The Reflection Tool integrates the task objectives and the knowledge returned by relevant tools to analyze whether the input text can accurately fulfill the task objective. The specific design of the prompt is illustrated in Figure \ref{fig:prompt_reflection}, where \texttt{\{Goals\}} represents our task objectives, \texttt{\{Relevant\_Tool\_Response\}} denotes the relevant knowledge returned by previously similar tools, and \texttt{\{Input\_Text\}} is the text that requires reflection in conjunction with the task objectives and related knowledge. Additionally, the prompt emphasizes that the output must align with the format of \texttt{\{Input\_Text\}}.


\section{Comparison with Advanced LLMs}\label{sec:reasonexp}

We conduct a comprehensive comparison between our agent and state-of-the-art LLMs (including reasoning models). To further evaluate the performance of our agent, we also enhance the baseline LLMs by constructing few-shot prompts, which are designed to improve their effectiveness. 

We selected two LLMs as the foundational models for our agent: GPT-4o-mini and GPT-4o. The versions used are consistent with those specified in Section \ref{sec:baseline}. For the evaluation, we randomly selected 10\% of the data for evaluation. 

As shown in Table \ref{tab:reasoningresult}, our agent achieves optimal results across all tasks. Specifically, the agent based on GPT-4 demonstrates outstanding performance in evidence reasoning (Task 2), indicating that models with larger parameter scales excel in tasks requiring logical reasoning and comprehension. However, in generative tasks such as fact probandum generation (Task 1), the agent based on GPT-4o-mini outperforms its got-4o counterpart, suggesting that smaller models may exhibit advantages in certain generation-oriented scenarios.

Furthermore, the experimental results reveal that the reasoning models (o1 and r1) outperform their base models (4o and v3) in Task 2, highlighting their enhanced capability in reasoning-intensive tasks. Conversely, in tasks more focused on generation (Task 1 and Task 3), the reasoning models underperform compared to their base models (4o and v3). This contrast underscores the importance of model architecture and scale in task-specific performance, particularly in balancing reasoning and generative capabilities.

Additionally, introducing examples to construct few-shot prompts effectively improves the model's performance on our benchmark. This demonstrates the utility of few-shot learning in enhancing task-specific adaptability and overall effectiveness.

\begin{table*}[htp]
\centering
\resizebox{1.0\textwidth}{!}{%
\begin{tabular}{l|ccc|ccc|ccc|c}
\hline \toprule
\multirow{2}*{\textbf{Approach}} & \multicolumn{3}{c|}{\textbf{Task I}} & \multicolumn{3}{c|}{\textbf{Task II}} & \multicolumn{3}{c|}{\textbf{Task III}} & \textbf{All} \\ 
\cline{2-11}
& $S_{fact}$-\textit{1} & $S_{fact}$-\textit{2} & $S_{fact}$-\textit{l} &$Pre$ & $Rec$ & $F_{evid}$ & $S_{exp}$-\textit{1} & $S_{exp}$-\textit{2} & $S_{exp}$-\textit{l} & $S_{c}$ \\ 
\hline
Claude3.5-sonnet-10&33.97&20.08&29.85&1.6&3.19&2.13&17.59&1.34&13.93&24.09
 \\
Claude3.5-sonnet-10 3 shot &36.69&20.86&32.25&2.87&5.64&3.8&17.77&2.63&12.54&27.03 \\
\hline
ChatGPT-4o&31.93&18.32&28.43&3.65&8.51&5.12&27.37&8.08&21.99&32.52 \\
ChatGPT-4o 3 shot &36.26&21.26&31.36&10.74&19.48&13.85&26.56&7.83&20.45&36.52 \\
\hline
ChatGPT-o1&30.56&17.63&27.61&7.54&23.84&11.45&23.56&5.45&17.63&32.25\\
ChatGPT-o1 3 shot &34.7&19.18&29.43&11.97&32.3&17.48&25.71&6.96&19.94&35.46 \\	

\hline
Deepseek-V3&30.23&15.12&26.13&7.2&14.35&9.59&29.23&9.54&23.97&32.62 \\
Deepseek-V3 3 shot &31.3&15.69&26.48&9.86&17.86&13.48&29.78&9.86&24.46&34.57 \\
\hline
Deepseek-R1 &29.74&12.3&25.46&10.78&21.25&14.3&21.58&2.94&16.53&32.38 \\
Deepseek-R1 3 shot &31.94&14.55&27.28&14.34&27.43&18.83&22.5&3.7&17.39&36.01 \\
\hline
\hline 
\task Agent (4o-mini)&\textbf{37.92}&\textbf{21.60}&\textbf{33.83}&8.38&39.48&13.83&31.92&9.32&25.49&\textbf{36.62} \\
\task Agent (4o) &34.88&21.37&29.86&\textbf{16.61}&\textbf{25.38}&\textbf{20.08}&\textbf{32.87}&\textbf{10.94}&\textbf{25.98}&36.11 \\
\bottomrule
\end{tabular}%
}
\caption{The results of advanced model}
\label{tab:reasoningresult}
\vspace{-3mm}
\end{table*}

\section{Ablation Study}\label{sec:ablation_study}

\begin{table*}[htp]
\centering
\resizebox{1.0\textwidth}{!}{%
\begin{tabular}{l|ccc|ccc|ccc}
\hline \toprule
\multirow{2}*{\textbf{Approach}} & \multicolumn{3}{c|}{\textbf{Task I}} & \multicolumn{3}{c|}{\textbf{Task II}} & \multicolumn{3}{c}{\textbf{Task III}} \\ 
\cline{2-10}
& $S_{fact}$-\textit{1} & $S_{fact}$-\textit{2} & $S_{fact}$-\textit{l} &$Pre$ & $Rec$ & $F_{evid}$ & $S_{exp}$-\textit{1} & $S_{exp}$-\textit{2} & $S_{exp}$-\textit{l} \\ 
\hline
\task Agent &37.92&21.60&33.83&8.38&39.48&13.83&31.92&9.32&25.49  \\ 
\midrule
\midrule
 - Pattern Match &37.65&21.45&33.53&7.65&38.44&12.76&31.33&9.46&24.45 \\
 - Multirole Checker &36.83&20.55&32.12&6.23&37.67&10.69&31.94&10.03&25.33\\
 - Legal Knowledge&35.70&19.56&31.98&8.35&37.23&13.64&30.56&8.68&23.80 \\
 - Knowledge Search &35.65&19.23&31.79&6.93&38.12&11.72&-&-&- \\
 - Emotion Check &37.12&20.85&32.95&-&-&-&-&-&- \\
\bottomrule
\end{tabular}%
}
\caption{Ablation Study.}
\label{tab:ablationresult}
\vspace{-3mm}
\end{table*}

In this section, we randomly selected 10\% of the data from the test set for evaluation to the impact of removing different tools from the toolkit on the agent's performance. The base model of the \task agent is GPT-4o-mini. Additionally, since the Fact Finding Head and Reflection Tool serve as the core tools and foundational experimental tools for the agent, ensuring the formatted output of the agent's results, the absence of these two tools would prevent the agent from producing formatted JSON data necessary for experimental validation. Therefore, we did not conduct ablation studies on these two tools.

As shown in Table \ref{tab:ablationresult}, the results of the ablation study indicate that each tool in our agent contributes positively to the task outcomes. Specifically, we observed that removing knowledge-based tools (Legal Knowledge and Knowledge Search) led to a significant decline in performance across all tasks, suggesting that the incorporation of domain-specific legal knowledge through these tools effectively enhances the agent's task execution. Furthermore, we found that removing the Multirole Checker tool also resulted in a noticeable decrease in the agent's task performance, indicating that the inclusion of multi-role judgment significantly improves the accuracy of task results.

\section{Future Work}
Despite our first try, challenges persist in AI adoption within the legal domain. Issues such as data privacy, imperceptible bias, the interpretability of AI enhancement, and the impact on traditional legal practices warrant further investigation. Future research directions involve addressing these challenges, enhancing interpretability and fostering interdisciplinary collaborations between AI agents and legal professionals.

\section{Related Work}\label{sec:related_work}
In the realm of judicial proceedings, the process can often be categorized into two fundamental phases~\citet{duxbury1995patterns,merryman2018civil}: (1) \textbf{Law Reasoning}, involving the determination of factual circumstances within a case; and (2) the \textbf{Law Application}, which pertains to the utilization and application of relevant legal statutes and principles to those identified facts.
Thus in this section, we review the current state of AI technology in these two subfields.

\subsection{AI for Law Application}
Law application refers to the process of applying the law. This involves determining the circumstances of the case, selecting the appropriate legal norm to be applied, interpreting the meaning of the chosen legal norm, and issuing a document that applies the legal norm to the relevant person or organization\footnote{\url{https://encyclopedia2.thefreedictionary.com/Application+of+Law}}.

Applying automated techniques to address a legal issue has a rich history. It can be traced back to early systems based on mathematics, such as~\citet{kort1957predicting,keown1980mathematical,lauderdale2012supreme}, which focus on analyzing cases using mathematical tools. Besides that, there are two main categories of AI approaches applied to Law: logic-based and data-driven approaches. The logic-based approach was introduced by~\citep{allen2013symbolic}, with its first appearance dating back to the 1980s. Around the same time, data-driven approaches were demonstrated by the HYPO system~\citep{ashley1991reasoning}. Some research has concentrated on addressing logical issues in legal documents and aims to clarify legal terminology, thereby contributing to the field of logic and the interactive representation of legal texts~\citep{branting2017data}. Additionally, a comprehensive description logic framework, proposed by~\citet{francesconi2014description}, builds upon Hohfeldian relations and effectively represents legal norms. Given that new laws are constantly introduced and existing laws are modified, the legal system is inherently dynamic, necessitating the establishment of an adaptable and modifiable model. To address this, the extension of defeasible logic has been widely employed and has yielded promising results~\cite{governatori2010changing,governatori2007variants,governatori2005temporalised}.
 Another significant challenge in the legal domain is analyzing vast repositories of case law, which poses an obstacle for legal professionals. Data-driven approaches are well-suited to tackle this issue. These approaches, employing techniques such as text mining~\citep{avgerinos2021legal}, information retrieval~\citep{sansone2022legal}, and semantic analysis~\citep{merchant2018nlp}, strive to extract valuable insights from legal texts and make judgments.
Furthermore, the rise of conversational AI technologies has resulted in the creation of legal chatbots~\citep{queudot2020improving}, which provide users with the ability to access legal information, receive help in completing legal documents, and receive guidance throughout legal processes through natural language interactions. In recent years, large language models (\textbf{LLMs}) have emerged as powerful tools, with several models specifically tailored to the legal domain being proposed~\citep{chalkidis-etal-2020-legal, song2023lawgpt, yuan2023chatlaw, colombo2024saullm7bpioneeringlargelanguage,fei2024internlmlawopensourcechinese, LexiLaw}. Concurrently, there has been a surge in the exploration of legal applications, including legal text comprehension and generation, legal documents analysis, legal violations detection, legal judgment predictions and etc.~\citep{huang2020generating, xu-etal-2020-distinguish,gan-etal-2023-exploiting, wu2023precedentenhanced,roegiest-etal-2023-questions, bernsohn-etal-2024-legallens, cao-etal-2024-pilot}. Still, these applications can only offer assistance to judges in duanting legal tasks but cannot delve deeply into the core of the judical process, which is rigorously revealing the ultimate fact through law reasoning.

\subsection{AI for Law Reasoning}
AI technology for law reasoning is relatively less common compared to law application, due to the complexity of structured law reasoning information. Structured facts exist in unstructured formats; teaching AI systems to extract accurate structured information at various levels is a complex task that requires sophisticated algorithms. Neglecting the structured law reasoning stage and directly employing the law application may lead to regulatory and compliance issues. Moreover, making all the facts, and shreds of evidence, as well as the reasoning process visible, makes AI applications more reliable.  

Compared to AI technology for law reasoning, the legal field has explored the simulation of evidential reasoning. As early as the late 1980s, Anne Gardner applied artificial intelligence techniques to the legal reasoning process and proposed a law reasoning program in Chapters 6\&7 of~\citep{gardner1987artificial}. In 1997, the book of~\citep{prakken1997logical} edited by Henry Prakken, provided a systematic account of how logical models of legal argumentation operate in legal reasoning. The book discusses various models of legal reasoning, including dialogical models, and provides a detailed analysis of the operation of different non-monotonic logics (such as logic programs, qualified reasoning, default logics, self-cognitive logics, etc.) in legal reasoning. According to Henry Prakken, these logics are of great importance for the development of artificially intelligent legal systems.
In 2009, the book of~\citep{walton2009hendrik}, edited by Henrik Kaptein, Henry Prakken, and Bart Verheij, proposed three practical ways of evidentiary reasoning: the statistical approaches, the storytelling-based approaches, and the argumentative approaches. Of these, the exploration of evidential reasoning with AI focuses on the latter two. Chapters 2\&3 illustrate the statistical approaches, Chapters 4\&5\&6 describe the storytelling-based approaches, Chapters 7\&8 compares storytelling-based approaches and argumentative approaches, and eventually, Chapters 9\&10 systematically describe the argumentative approaches. Floris J. Bex~\citep{bex2011arguments} attempt to construct a hybrid theory of evidence based on argumentation and explanation, raises the issue of legal reasoning in AI. All the above literature discusses the theoretical possibilities of combining AI technology with legal reasoning theory, and suggests that the analysis of these argumentation patterns could be the logical architecture of an AI legal system. Till 2012, Ephraim Nissan (especially Volume 1, Chapter 3 of~\citep{nissan2012computer}) attempts to introduce the Wigmore diagram and the Turmin logical argumentation model into computer programs, and attempts to place them under the calculation of certain AI systems. Then in 2017, Floris J. Bex and Serena Villata~\citep{wyner2017legal} introduced and summarized the application of AI technology in the field of legal systems, especially the integration of AI technology and legal argumentation.
It is evident that the ongoing research on legal systems for artificial intelligence has now reached the stage of developing models for evidence reasoning.

\section{Legal Cases Examples}\label{sec:case_examples}
To highlight the significance of law reasoning, we provide examples that are widely recognized where different judicial facts, evidence, and experience have impacted different results. Recognizing these instances is crucial for maintaining public trust. We provide an notable example in  Figure \ref{fig:motivation}.
The \textit{Rex v. Bywaters and Thompson} is one of England’s most famous causes~\citep{anderson2005analysis}.
The \textit{Rex v. Bywaters and Thompson} is one of England’s most famous causes c´el`ebres~\citep{anderson2005analysis}. The case is an example of rough as well as speedy ``justice.''
On January 9, 1923, Frederick Bywaters and Edith Thompson were hanged for the murder of Edith’s husband Percy, just three months and six days after his death. Public opinion at the time and subsequent commentators have been divided on the question of whether Edith had instructed her mistress Bywaters to kill her husband.
 All evidence indicates that Edith's marriage with her husband was unhappy. 
 They met a young steward named Bywater on a cruise ship by chance. After that, they exchanged letters frequently. Although some of the letters were burned, 
 Edith met a young steward named Bywater on a cruise ship by chance. All evidence indicates that Edith and Bywater had already reached the most reprehensible intimate relationship. Before Bywater killed Edith's husband, although there is no direct evidence that Edith instigated Bywater to kill her husband, due to the possibility of their meeting before the crime and the evidence of their shared interests, with the usual experience of ``the elderly tend to dominate'', Edith was sentenced to death.
However, some commentators think that if we consider that Edith is an older woman, according to common sense, she would often take on the role of a mother, and because she is a woman who is less inclined to dominate others, she would not make such an inciting behavior.

Similar to the case of ``Rex v. Bywaters and Thompson'', in the case of ``Nanjing Pengyu''\footnote{\url{https://www.chinacourt.org/article/detail/2014/07/id/1352051.shtml}}, the judge applied the wrong experience that a person would not help someone who had fallen, but only the person who caused the fall. This erroneous experience resulted in a miscarriage of justice.

\section{Dataset Reliability Analysis}\label{sec:data_reliability_analysis}
We assess the reliability of our new dataset through manual review. 
Human workers were enlisted to assess whether the labeled data is aligned with the requirements and adhered to the schema. A multiple-choice questionnaire was created, consisting of fifteen labeled samples for scoring. The questionnaire included three possible responses: correct, neutral, and incorrect. Workers were asked to indicate whether the labeled samples were correct. Each participant received a compensation of approximately \$8.5 per hour. The results showed that the majority of the workers found the labeled samples to be correct (ratio of correct 95\%). This indicates that the labeled data aligns with the requirements and adheres to the schema.









\end{document}

%% file: config.tex
\usepackage{graphicx}
\usepackage{amsmath,amssymb,amsthm,amsfonts}
\usepackage{bbm}
\usepackage{acronym}
\usepackage{enumitem}
\usepackage{balance}
\usepackage{xspace}
\usepackage[skip=3pt]{subcaption}
\usepackage[skip=3pt]{caption}
\usepackage[switch]{lineno}
\usepackage{url}
\usepackage{multirow}
\usepackage{booktabs}
\usepackage{ragged2e} 
\usepackage{mathrsfs}
\usepackage{arydshln}
\usepackage{array}
\usepackage{colortbl}
\usepackage{pifont}

\definecolor{myblue}{RGB}{0,105,252}
\definecolor{myred}{RGB}{247,96,102}
\definecolor{mygray}{gray}{0.4}

\makeatletter
\DeclareRobustCommand\onedot{\futurelet\@let@token\@onedot}
\def\@onedot{\ifx\@let@token.\else.\null\fi\xspace}

\makeatother

\acrodef{nlp}[NLP]{natural language processing}
\acrodef{plm}[PLM]{pretrained language model}
\acrodef{sota}[SOTA]{state-of-the-art}
\acrodef{bs}[BS]{Beam Search}
\acrodef{mhs}[MHS]{Metropolis-Hastings Sampling}
\acrodef{hs}[HS]{Hybrid Search}
\acrodef{uas}[UAS]{unlabeled attachment score}
\acrodef{dda}[DDA]{Directed Dependency Accuracy}
\acrodef{sota}[SOTA]{state-of-the-art}
\acrodef{pos}[POS]{part-of-speech}
\acrodef{asr}[ASR]{attacking success rate}
\acrodef{ppl}[PPL]{Perplexity score}
\acrodef{cqr}[CQR]{Conversational Question Reformulation}
\acrodef{cqa}[CQA]{Conversational Question Answering}
\acrodef{mcqr}[MTCQR]{Multi-Topic Conversational Question Reformulation}
\acrodef{amt}[AMT]{Amazon Mechanical Turk}
\acrodef{mtcl}[MTCL]{Multi-Topic Contrastive Learning}